\documentclass[lettersize,journal]{IEEEtran}
\usepackage[table]{xcolor}
\usepackage{amsmath,amsfonts}
\usepackage{algorithmic}
\usepackage{algorithm}
\usepackage{array}
\usepackage[caption=false,font=normalsize,labelfont=sf,textfont=sf]{subfig}
\usepackage{textcomp}
\usepackage{stfloats}
\usepackage{url}
\usepackage{verbatim}
\usepackage{graphicx}
\usepackage{cite}
\usepackage{tikz}
\usepackage{makecell}
\usepackage{multirow}
\usepackage{booktabs}
\usepackage{threeparttable}
\usepackage{color}
\usepackage{upgreek}
\usepackage{utfsym}
\newcommand{\etal}{{et al.\,}}
\usepackage{hyperref}

\usepackage{orcidlink}
\hypersetup{hidelinks}

\begin{document}

\title{BiFingerPose: Bimodal Finger Pose Estimation for Touch Devices}

\author{Xiongjun Guan$^{\orcidlink{0000-0001-8887-3735}}$, 
	Zhiyu Pan$^{\orcidlink{0009-0000-6721-4482}}$,
	Jianjiang Feng$^{\orcidlink{0000-0003-4971-6707}}$, ~\IEEEmembership{Member, IEEE}, 
	and Jie Zhou$^{\orcidlink{0000-0001-7701-234X}}$, ~\IEEEmembership{Fellow, IEEE}
	
	\thanks{
		This work was supported in part by the National Natural Science Foundation of China under Grant 62376132 and 62321005. (\emph{Corresponding author: Jianjiang Feng}.)}
	\IEEEcompsocitemizethanks{
    \IEEEcompsocthanksitem
	The authors are with Department of Automation, Tsinghua University, Beijing 100084, China (e-mail: \url{gxj21@mails.tsinghua.edu.cn}; \url{pzy20@mails.tsinghua.edu.cn}; \url{jfeng@tsinghua.edu.cn}; \url{jzhou@tsinghua.edu.cn}).
	}
	
}

\markboth{IEEE TRANSACTIONS ON MOBILE COMPUTING, ~Vol.~x, No.~x, June~2025}%
{Shell \MakeLowercase{\textit{et al.}}: A Sample Article Using IEEEtran.cls for IEEE Journals}


\maketitle

\begin{abstract}
Finger pose offers promising opportunities to expand human computer interaction capability of touchscreen devices. 
Existing finger pose estimation algorithms that can be implemented in portable devices predominantly rely on capacitive images, which are currently limited to estimating pitch and yaw angles and exhibit reduced accuracy when processing large-angle inputs (especially when it is greater than 45 degrees).
In this paper, we propose BiFingerPose, a novel bimodal based finger pose estimation algorithm capable of simultaneously and accurately predicting comprehensive finger pose information.
A bimodal input is explored, including a capacitive image and a fingerprint patch obtained from the touchscreen with an under-screen fingerprint sensor.
Our approach leads to reliable estimation of roll angle, which is not achievable using only a single modality.
In addition, the prediction performance of other pose parameters has also been greatly improved.
The evaluation of a 12-person user study on continuous and discrete interaction tasks further validated the advantages of our approach.
Specifically, BiFingerPose outperforms previous SOTA methods with over 21\% improvement in prediction performance, $2.5\times$ higher task completion efficiency, and 23\% better user operation accuracy, demonstrating its practical superiority.
Finally, we delineate the application space of finger pose with respect to enhancing authentication security and improving interactive experiences, and develop corresponding prototypes to showcase the interaction potential.
Our code will be available at https://github.com/XiongjunGuan/DualFingerPose.
\end{abstract}

\begin{IEEEkeywords}
Finger pose, orientation, touch, interactive surface, fingerprint, capacitive sensing, smartphone.
\end{IEEEkeywords}

\section{Introduction}
\IEEEPARstart{T}{ouch} based interaction has gained widespread adoption in modern consumer electronic devices, including smartphones, tablets, and smartwatches, owing to its intuitiveness and reliability. 
However, current commercial touch devices only capture low-resolution capacitive images of the fingers and determine 2D contact positions \cite{nam2021review}. 
This limitation results in the oversight of numerous freedoms inherent in dexterous finger movements, consequently restricting the advancement of interaction operability and flexibility.
To expand the input vocabulary of touch interaction, researchers have investigated various finger state measurements, including gestures \cite{kong2021continuous,tan2022enabling,kong2024FedAWR}, physical vibrations \cite{yang2022enabling,jiang2024two}, acoustic feedback \cite{wu2021echowrite,ling2022UltraGesture,zhou2024fingerpattern}, pressing duration \cite{elias2010multitouch}, applied pressure \cite{boceck2019force}, shear force \cite{yu2023PrintShear}, finger shape \cite{gil2017TriTap}, contact area \cite{ikematsu2020ScraTouch,wu2024all}, behavior signature \cite{ren2020signature} and finger pose \cite{holz2010generalized, zaliva2012finger,xiao2015estimating,mayer2017estimating,duan2023finger,he2024trackpose}.

Among these explored extensions of touch interaction, the application of finger pose has demonstrated remarkable advantages in accuracy, efficiency, and comprehensibility.
Compared to other finger properties, such as contact position, contact area, and finger shape, finger pose offers higher degrees of freedom and a larger range.
This versatility makes it widely applicable across a diverse array of manipulation, selection, adjustment, and other high-level tasks \cite{vogelsang2021design}.
In addition, finger pose interactions are very common and frequently practiced in daily life, making them easier to perceive and control.
Moreover, they exhibit great compatibility and complementarity with common actions such as clicking, sliding, and long pressing, and can be integrated to further enhance interactive applications. 
It is worth mentioning that this convenience and richness demonstrate a particularly great potential for portable devices with limited area or situations with obstruction \cite{gil2018fingers,vogelsang2021design}. 
In addition to serving as input, accurate pose estimation can also provide assistance for other interaction designs, such as correcting touch points by referencing the finger angle during touch \cite{holz2010generalized} or eliminating accidental touches by checking extreme finger touch angles \cite{goguey2018characterizing}.

Finger pose estimation shows great potential in touch interaction.
However, \textbf{capacitive image} based solutions in previous research primarily concentrated on 3D pose and were limited to estimating only two angles: pitch and yaw \cite{xiao2015estimating,mayer2017estimating,Ullerich2023ThumbPitch,he2024trackpose}.
Moreover, their prediction performance typically declines significantly when dealing with substantial touch angles \cite{Ullerich2023ThumbPitch,he2024trackpose}.
On the other hand, the fingerprint modality offers higher resolution, and related studies on \textbf{full fingerprints} generally enable more accurate estimation of all 2D \cite{yin2021joint,he2022PFVNet,duan2023estimating} or 3D \cite{duan2022estimating,duan2023finger} finger pose parameters.
Nevertheless, it necessitates large area and high spatial resolution (500 ppi, $512\times512$ pixels) fingerprint sensors, which are impractical for nowadays portable electronic devices such as mobile phones, primarily due to their significant size and cost constraints.
At present, the popular solution is to deploy a compact fingerprint sensor (about 500 ppi, $120\times120$ pixels) beneath the touch screen, which can simultaneously capture both \textbf{capacitive image} and \textbf{fingerprint patch} upon touch \cite{koundinya2014support, peng2021under, yin2021optical}.
A considerable number of mobile phones have already adopted this compromise deployment plan \cite{market2024indisplay,market2024under,qualcomm}.
However, a significant reduction in the effective collection area severely diminishes fingerprint based pose estimation accuracy \cite{duan2023finger, guan2025joint}.

\begin{figure*}[!t]
    \centering
    \includegraphics[width=.9\linewidth]{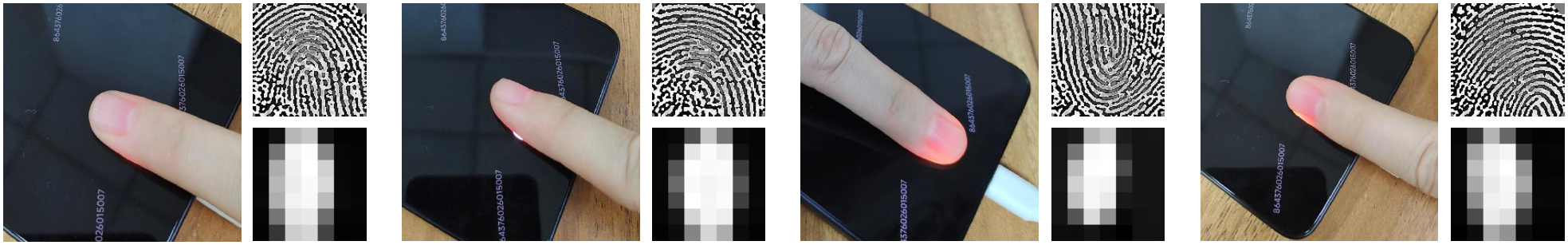}
    \vspace{-2mm}
    \caption{Examples of capacitive images and fingerprint patches in different touch poses.}
    \label{fig:intro_data}
\end{figure*}

Figure \ref{fig:intro_data} illustrates four examples captured by a smartphone equipped with the above sensor solution under different finger poses. 
It can be observed that relying solely on capacitive images (which provide global, coarse contour information) or partial fingerprint patches (offering local, fine texture details) for 2D or 3D finger pose estimation is a challenging task. However, by integrating data from both modalities, we gain access to more comprehensive information, enabling robust pose inference.
This inherent information advantage of individual modalities, combined with their simultaneous acquisition capability in existing portable electronic devices, presents a compelling opportunity to leverage their complementarity for more stable and accurate prediction of complete finger pose information from touch interactions.

\begin{figure*}[!t]
\centering
  \includegraphics[width=.85\textwidth]{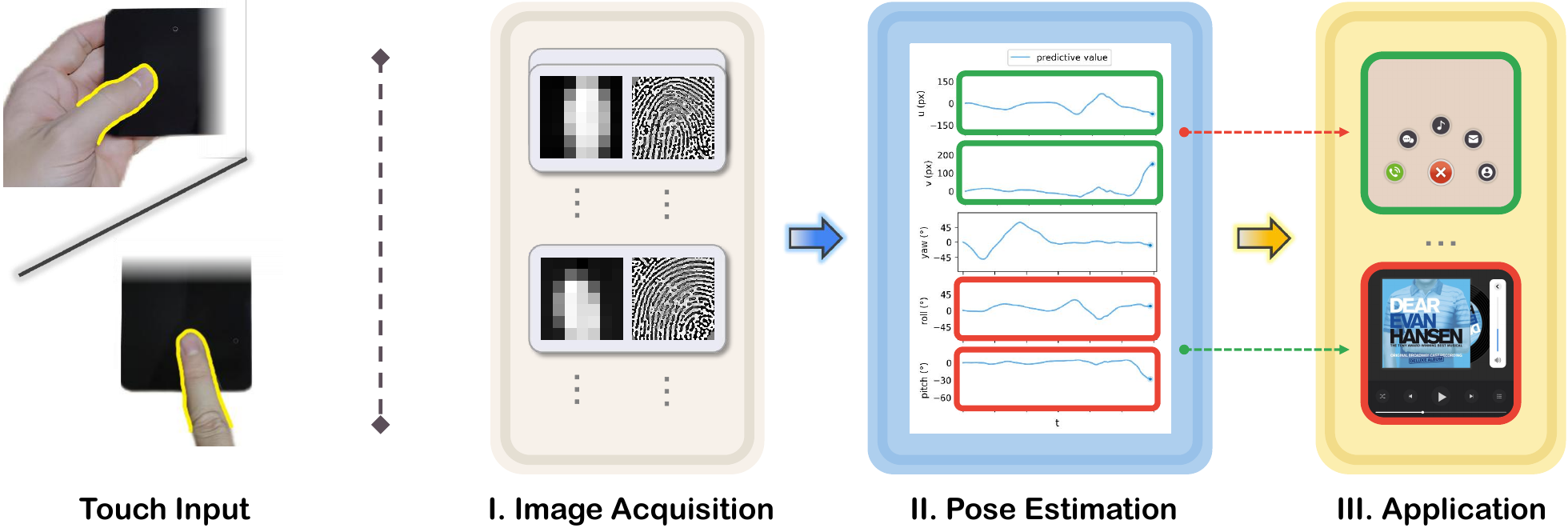}
  \caption{Users change their finger pose in any comfortable way to perform interactive operations on touch devices.
  Our BiFingerPose ultilizes capacitive image and fingerprint patch captured by touchscreen devices with under-screen fingerprint sensors to provide robust, precise and comprehensive finger pose estimation, which can be used for various applications. 
  Notably, our solution eliminates the need for extra devices or the storage of pre-registered fingerprint information, ensuring both user privacy and a highly convenient experience.
  }
  \label{fig:teaser}
\end{figure*}

In this paper, we propose a \textbf{Bi}modal based \textbf{Finger} \textbf{Pose} estimation algorithm, termed \textbf{BiFingerPose}, which achieves precise and stable predictions for all degrees of freedom in 2D/3D pose parameters.
Figure \ref{fig:teaser} shows the complete application process of our BiFingerPose.
Concretely, we first designed a convolutional neural network to predict the 2D finger pose, and subsequently mapped it to the 3D pose using our proposed pose transformation function.
Unlike previous regression-based pose estimation methods \cite{mayer2017estimating,yin2021joint,duan2023finger,Ullerich2023ThumbPitch,he2024trackpose}, we introduce trigonometric probability distribution vectors to assist our network in better understanding the adjacency relationships between poses.
On the other hand, the capacitive image and fingerprint patch captured from the touch screen (equipped with an under-screen fingerprint sensor) are utilized as bimodal inputs, instead of relying on a single modality as in previous approaches \cite{mayer2017estimating,yin2021joint,duan2023finger,duan2023estimating,he2024trackpose}.
By leveraging the complementarity of these two inputs, the roll angle can be predicted with relatively high accuracy.
Extensive experiments and user studies demonstrate that our method achieves state-of-the-art (SOTA) performance on mainstream finger modalities.
Specifically, compared to related advanced methods, BiFingerPose achieves a remarkable improvement of over 21\% in pose prediction accuracy, while demonstrating 2.5 times greater task efficiency and a 23\% enhancement in user operation precision.
We also conduct a discussion of the potential applications of 2D and 3D finger pose estimation. 
Several application prototypes are provided to qualitatively illustrate their practical utility in user interaction.

It should be emphasized that our solution lies in its data-driven foundation, which fundamentally departs from traditional template-matching approaches. 
This design eliminates the necessity for pre-registered fingerprint templates \cite{holz2010generalized,duan2022estimating,liu2023printype}, thereby significantly enhancing user privacy and streamlining the user experience.
Furthermore, a distinct advantage of BigFingerPose, particularly when compared to existing solutions\cite{watanabe2012contact,streli2022TapType,zhang2022typeanywhere,liang2021DualRing}, is its ability to operate without requiring any hardware modifications. This enables its seamless integration into existing touch devices equipped with under-display fingerprint sensors, solely through software updates.
Moreover, while existing methods exhibit significant estimation errors when the yaw angle exceeds $45^\circ$ \cite{yin2021joint,duan2023estimating,xiao2015estimating,mayer2017estimating,Ullerich2023ThumbPitch,he2024trackpose}, our approach maintains stable and accurate prediction across the full $360^\circ$ range. 
This breakthrough enables users to interact with the device in any gestures (such as vertical, horizontal, or inverted handheld) through diverse finger poses, providing smoother, more stable, and more reliable estimates in different usage environments, which is crucial for the interactive experience \cite{batmaz2019effect,batmaz2020no}.
 
In summary, the main contributions of this paper are as follows:
\begin{itemize}
	\item We propose a novel framework for finger pose estimation, called BiFingerPose. 
    A bimodal based approach combining capacitive image and fingerprint patch is explored, which can significantly improve the accuracy and stability of pose estimation compared to previous single-modality approaches.
	\item We introduce a novel angle representation using trigonometric probability distribution vectors to provide superior optimization guidance for pose estimation neural networks.
	\item We demonstrate that the 2D finger pose can be mapped to the 3D pose with minimal error via simple polynomials.
    This conversion approach enables researchers to seamlessly adapt existing finger pose estimation algorithms, developed under one definition, to execute interactive tasks that may be more effective with other pose definitions.
	\item Extensive experiments and user studies are conducted to evaluate representative SOTA algorithms, demonstrating the superiority of our proposed method. Moreover, potential application scenarios of finger pose are discussed and demonstrated to qualitatively demonstrate its potential applications.
\end{itemize}

\section{Related Work}

A notable distinction in finger pose research lies in the differentiation of modalities.
Different types of sensors lead to pronounced modal variations in captured images of finger touches, resulting in significant differences in the form and completeness of touch information.
In addition, some studies suggest obtaining other auxiliary information, such as point clouds or infrared images, through additional sensors.
In this section, we first introduce the typical definition of finger pose, and then review existing works based on input modalities.
Moreover, we also provide information about underscreen fingerprint sensor technology for reference.
Finally, the technical solutions for feature fusion are briefly summarized and a basic overview is provided.

\begin{figure*}[!t]
    \centering
    \includegraphics[width=.65\linewidth]{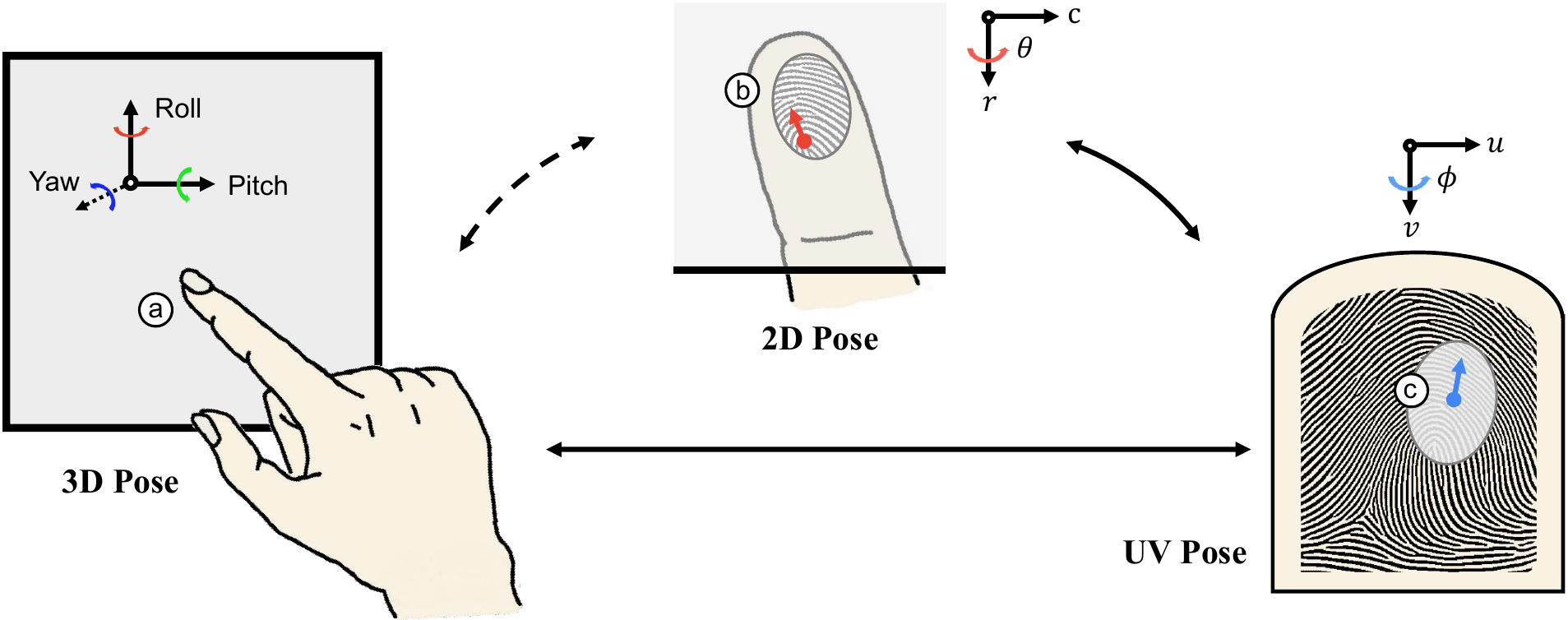}
    \vspace{-2mm}
    \caption{Definition and conversion process of three finger pose types.
        The precise mapping relationship between (a) 3D pose (roll, pitch, yaw of fingertip) and (b) 2D pose (position and angle of the fingerprint center in the screen coordinate system) can be established conveniently through (c) UV pose (position and angle of the contact center in the coordinate system of the normalized rolled fingerprint) introduced in this paper.}
    \label{fig:intro}
\end{figure*}

\subsection{Finger Pose Definition}
Figure \ref{fig:intro} illustrates the definitions of different forms of finger pose.
Researchers typically employ the 3D pose \cite{xiao2015estimating,mayer2017estimating}, which corresponds to the three spatial rotation angles of the fingers, for interactive control.
Furthermore, some researchers utilize the 2D pose \cite{yin2021joint,duan2023estimating}, as a representation, defined by the fingerprint center and a positive direction. 
In fact, these two postures are approximately equivalent. 
In this paper, we achieve the conversion of poses from 2D to 3D through the UV pose depicted in Figure \ref{fig:intro}, and demonstrate that this conversion is nearly lossless.
The UV pose coordinate system takes the center of the rolled fingerprint as the origin, and the positive direction of the ordinate axis is consistent with the positive direction of its 2D pose.
Through this conversion approach, researchers can effectively repurpose established finger pose estimation algorithms, originally formulated under a particular definition, to undertake interactive tasks that are more appropriately aligned with different pose definitions.

\subsection{High Resolution Image Based Methods}
Among different modals, fingerprint images contain the most complete and clear contour and texture information, which are typically collected using high resolution sensors (approximately 500 ppi) that can cover the entire fingertip.
Extensive studies have been conducted on their 2D pose estimation, which has been widely applied in person identification \cite{yin2021joint, duan2023estimating}.
However, the absence of a direct association between the 2D pose and the intricate 3D geometry of the finger constrains their application scope in interactive systems.
Holz \etal \cite{holz2010generalized} proposed an algorithm for calibrating touch positions using fingerprint images, which yields a by-product of the 3D finger pose.
A set of gallery fingerprints with corresponding finger angles are pre-registered and then matched with input fingerprints to predict 3D angles.
Duan \etal \cite{duan2022estimating} reconstructed a 3D finger surface from fingerprint sequence frames with their angles during a registration stage and estimated the 3D angle by solving projection parameters based on keypoint matching during the matching phase.
In addition, some recent studies \cite{he2022estimating,duan2023finger} directly predict 3D poses based on deep neural networks, further improving the accuracy and speed.
However, large area fingerprint sensors are rare in portable touch devices.

On the other hand, fingerprint patches can be considered as a special case of fingerprints, commonly obtained on mobile devices with limited sensing areas.
Due to severe information loss, current researches still have large pose estimation errors and are difficult to support practical applications \cite{he2022PFVNet}.

\subsection{Low Resolution Image Based Methods}
In contrast to high resolution fingerprints, where ridges are clearly visible, capacitive images typically have a resolution of only around 10 ppi, and the touch state is primarily inferred from the contour and capacitance value.
Due to their widespread use on mobile devices, capacitive images have become the most commonly used input modality for interactive surface-related works.
Zaliva \etal \cite{zaliva2012finger} defined multiple sets of characteristics of the foreground region, such as centroids, average intensity, and area, to estimate 3D finger angles and also used them for gesture recognition.
Xiao \etal \cite{xiao2015estimating} hypothesized that the contact area has a "comet" shape.
On this basis, they defined 42 features and estimated the pitch and yaw angles of the finger using a Gaussian Regression Model.
With the development of deep learning technology, Mayer \etal \cite{mayer2017estimating} used convolutional neural networks to directly regress pitch and yaw angles from capacitance images, achieving higher accuracy than previous empirically designed algorithms.
Ullerich \etal \cite{Ullerich2023ThumbPitch} used a lightweight network to estimate the pitch angle and explored its potential application in one-handed interaction scenarios.
He \etal \cite{he2024trackpose} introduced a self-attention mechanism in their network to fuse multi-frame features, further improving the accuracy of yaw and pitch angle estimation.
The roll angle is usually ignored in these studies because distinguishing it from relatively symmetric low-resolution images can easily cause confusion, although rolling the finger is a very effective signal while easy to implement \cite{roudaut2009MicroRolls}.
Furthermore, the current prediction accuracy and stability is still unsatisfactory, especially in the case of large angle input \cite{Ullerich2023ThumbPitch,he2024trackpose}, which hinders its popularization in daily applications \cite{vogelsang2021design}.
It is worth mentioning that Ahuja \etal \cite{ahuja2021TouchPose} also employed a network to estimate palm posture on large touch screens. 
While the combined information from fingers and palms achieves high accuracy, this method is less suitable for devices with limited touch areas, such as mobile phones or watches.

It is worth mentioning that mask-based pose estimation has not been systematically studied.
Wang \etal \cite{wang2009detecting} used an ellipse to approximate the contour shape of the touch region and estimated the yaw angle based on its long axis.
Dang \etal \cite{dang2011usage} extracted the outer contour of the fingers and also used ellipses to estimate the yaw angle.
Due to the lack of information, previous approaches \cite{wang2009detecting,dang2011usage} are confined to estimating only the angle of yaw and exhibit relatively high error rates.
Nevertheless, mask-based pose estimation is easy to deploy while ensuring that user identity information will never be leaked from the source, making it still a certain potential for application.
In addition, there are some inspiring works that can accurately infer high-resolution contact masks from capacitive images \cite{mayer2021super,streli2021CapContact,rusu2023deep}.
Overall, mask based methods require further investigation to rival fingerprint or capacitive image based solutions.

\subsection{Additional Sensor Based Methods}
In addition, some researchers proposed the introduction of additional sensors to gather more information.
Rogers \etal \cite{rogers2011anglepose} investigated the utilization of capacitive sensor arrays to acquire 2D contact positions and 3D finger angles.
Watanabe \etal \cite{watanabe2012contact} used additional cameras attached to the fingertip to monitor the intensity of reflected light.
Some solutions \cite{Kratz2013PointPose,mayer2017feasibility} used depth cameras to capture point clouds of the fingers.
Moreover, Liang \etal \cite{liang2021DualRing} directly utilized a ring-shaped IMU to capture the state and movement of the fingers.
In general, these approaches can achieve relatively stable performance, benefiting from direct monitoring of specific physical signals.
However, the requirement for additional sensors increases deployment costs and usage complexity, thereby limiting their applicability across a wide range of scenarios.
Taking portable electronic devices as an illustration, users typically desire to operate them without the need for additional physical components, thereby achieving a lightweight experience.

\subsection{Under Screen Fingerprint Sensor}
With the advancement of under-screen fingerprint technology, researchers have developed miniaturized high-resolution fingerprint sensors that can be integrated into practical devices to capture fingerprint images \cite{peng2021under, yin2021optical, koundinya2014support}, significantly enhancing the practicality of fingerprints in interactive applications.
Currently, a considerable number of devices already incorporate under-screen fingerprint sensors within the touch screen, allowing for simultaneous acquisition of capacitive images and fingerprint patch \cite{market2024indisplay,market2024under,qualcomm}.
Inspired by this, in this paper, we explore finger pose estimation based on this bimodal input to fully leverage the complementary advantages of the combined information.

\subsection{Feature Fusion}
To address the inherent limitations of individual sensors, information fusion technology has been developed. By integrating data from multiple competitive or complementary sensors, this approach yields environmental perception results that are more accurate, reliable, and comprehensive than those achievable with any single sensor \cite{yeong2021sensor}.
Based on the stage of data fusion, these solutions can be roughly divided into three categories:
(1) Early fusion (data-level fusion), which directly combines raw sensor data at the forefront of the data processing pipeline \cite{liu2013data};
(2) Intermediate fusion (feature-level fusion), which is currently the most active research field. 
Under this strategy, raw data from various sensors are first feature extracted through independent modules, and then these fused in the intermediate layer \cite{park2017rdfnet}.
(3) Late fusion (decision-level fusion). In this scheme, each sensor independently perceives and generates decision or prediction results, and then integrates these results through later modules to generate more reliable predictions \cite{teng2021multisensor}.
Given the inherent lack of a direct projection correlation between capacitance images and fingerprint blocks, we devised a hybrid fusion scheme based on (2) and (3). 
Our final approach employs Mixture of Experts (MOE) for feature-level processing of both independent and mixed features, and subsequently integrates their outputs at the decision level using probability distribution vectors.
This design aims to combine the advantages of different fusion stages, thereby enhancing the model's adaptability.

\section{Data Collection} \label{sec:data-collection}
Considering the current scarcity of publicly available large-scale databases with 2D/3D pose ground truth, our initial efforts are focused on organizing suitable data to facilitate the development and evaluation of our algorithms.
We first collected a large-scale database with simulated capacitive images, fingerprint patches, and 2D/3D poses, referred to as the \textbf{P}lain and \textbf{R}olled \textbf{F}ingerprint database (\emph{PRF}), which was utilized for training and testing.
Subsequently, we collected another real-world dataset of capacitive images, fingerprint patches, and 2D/3D poses, referred to as the \textbf{C}apacitive image and \textbf{F}ingerprint \textbf{P}atch dataset (\emph{CFP}), which was utilized for fine-tuning and testing.
The following text will sequentially introduce the equipment, participants, and data collection process.
It is worth noting that the additional equipment is solely required during the data collection phase. 
For practical applications, users simply need to update the software on devices capable of dual-modal image acquisition to utilize it.

\begin{figure*}[!t]
	\centering
	\subfloat[]{\includegraphics[height=3.5cm]{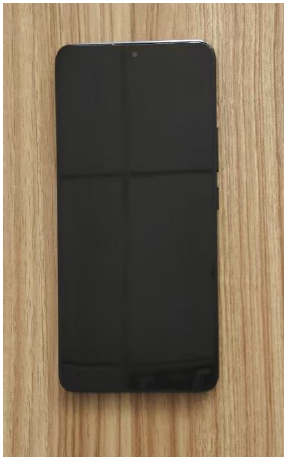}%
        \label{fig:db_ex1}
	\vspace{-1mm}}
	\hfil
	\subfloat[]{\includegraphics[height=3.5cm]{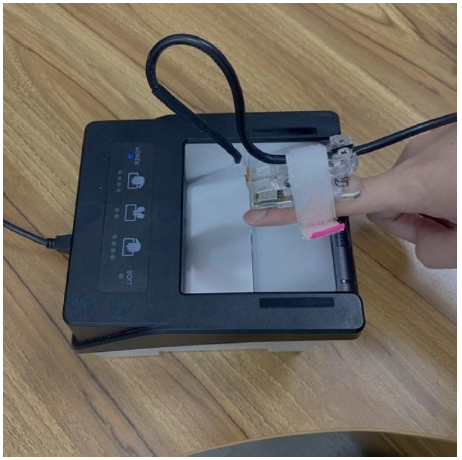}%
	\label{fig:db_ex2}
        \vspace{-1mm}}
	\hfil
	\subfloat[]{\includegraphics[height=3.5cm]{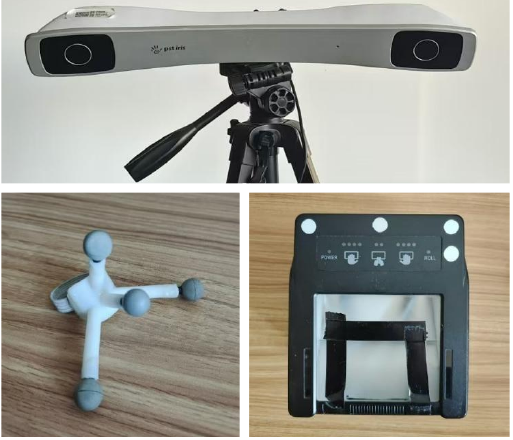}%
	\label{fig:db_ex3}
        \vspace{-1mm}}
	\hfil
	\subfloat[]{\includegraphics[height=3.5cm]{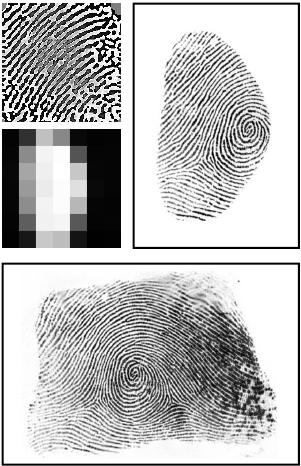}%
	\label{fig:db_ex4}
        \vspace{-1mm}}
	
	\caption{Smartphone (a), IMU (b) and optical tracker (c) based data acquisition system. 
    In particular, an optical fingerprint acquisition instrument is used to collect fingerprint images to obtain 2D poses in combination with fingerprint patches collected by a mobile phone.
    Samples of capacitive image \& fingerprint patch (from smartphone) and plain / rolled fingerprint (from optical scanner) are shown in (d).}
	\label{fig:db_ex}
\end{figure*}

\begin{figure}[!t]
\centering
	\subfloat{\includegraphics[width=.32\linewidth]{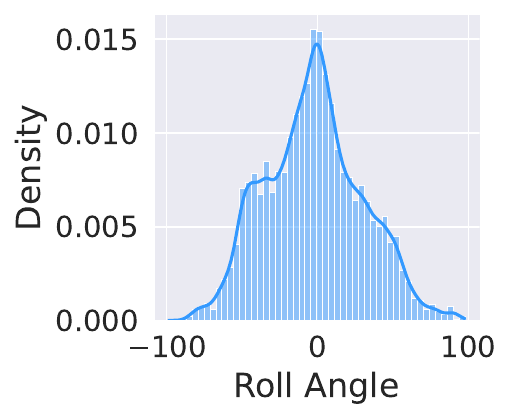}}
	\hfil
	\subfloat{\includegraphics[width=.32\linewidth]
    {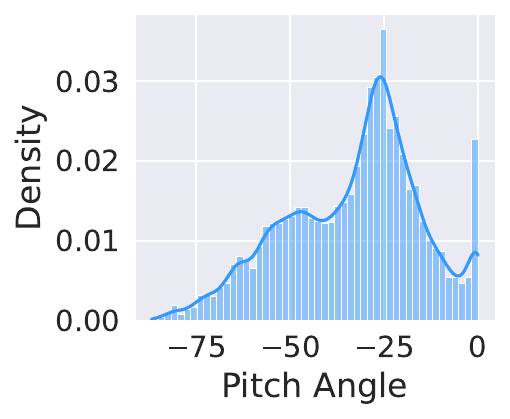}}
    \hfil
	\subfloat{\includegraphics[width=.32\linewidth]{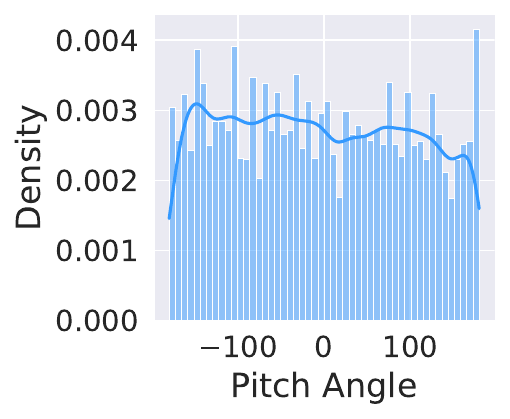}}
    \vspace{-2mm}
\caption{Distributions of roll, pitch and yaw angles in \emph{PRF}.}
\label{fig:db_distribution_RPF}
\end{figure}

\begin{figure}[!t]
	\centering
	\subfloat{\includegraphics[width=.32\linewidth]{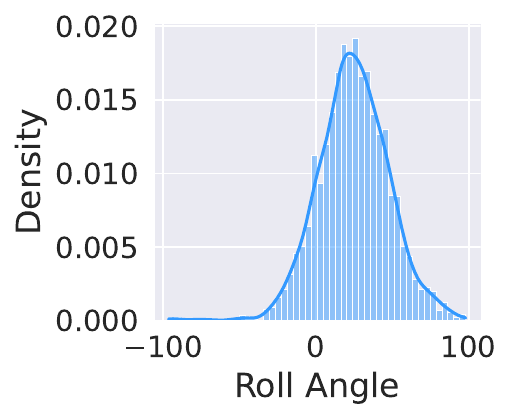}}
	\hfil
	\subfloat{\includegraphics[width=.32\linewidth]{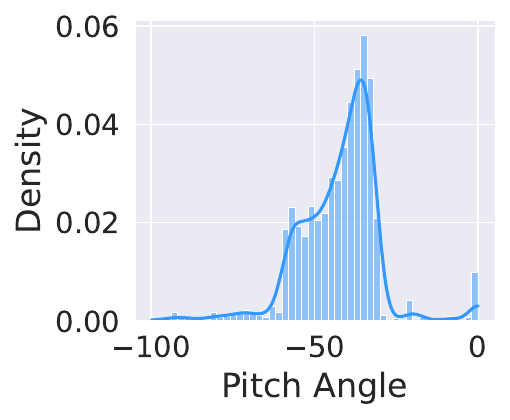}}
    \hfil
	\subfloat{\includegraphics[width=.32\linewidth]{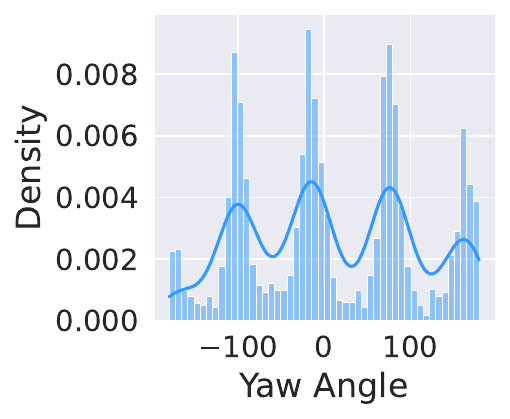}}
    \vspace{-2mm}
\caption{Distributions of roll, pitch and yaw angles in \emph{CFP}.}
\label{fig:db_distribution_CFP}
\end{figure}

\subsection{Apparatus} \label{subsec:data_collection-apparatus}

As shown in Figure \ref{fig:db_ex1},  we collected the bimodal data for the \emph{CFP} dataset using a Xiaomi 24069RA21C smartphone, which is equipped with an under-screen fingerprint sensor. 
This data comprises two types: capacitive images with a resolution of 10 ppi ($18\times40$ pixels) and a sampling rate exceeding 30 Hz, and fingerprint patches with a resolution of 500 ppi ($120\times120$ pixels) and a maximum sampling rate of 15 Hz.
The device is equipped with Snapdragon 8s Gen 3 Mobile Platfor and an integrated Adreno GPU.
The entire screen measures $160\times75 \:\text{mm}^2$, with the actual interactive input area being roughly $20\times20 \:\text{mm}^2$. 
This is the region equipped with the fingerprint sensor.
The device can simultaneously capture and return both modes of images during touch interaction at a frequency of 15 Hz.
Figure \ref{fig:db_ex4} shows examples of collected images.

As shown in Figure \ref{fig:db_ex2}, we employed an inertial motion unit (IMU) to collect a portion of the 3D pose in \emph{RPF}.
The device is mounted on a ring and can provide pose information at a frequency of 50 Hz.
We used an optical tracker to obtain another portion of 3D pose information from a ring equipped with 4 reflective balls and sticker markers on the scanner (as illustrated in Figure \ref{fig:db_ex3}) in \emph{RPF} and \emph{CFP}.
A binocular optical tracking camera was employed for landmark localization.
The pose definitions had been calibrated with the IMU in advance, ensuring that the 3D poses from two acquisition rounds could be considered consistent without deviation.
It should be noted that the utilization of both IMU and optical tracker arose from practical considerations pertaining to distinct data acquisition processes and evolving project timelines. 
Nevertheless, the label consistency is maintained as both systems are able to ensure the recording of finger pose with an error of less than $1^\circ$, and a common calibration standard was rigorously applied to ensure the uniformity of data annotation.
The calibration process involves recording a 'zero pose' after volunteers wear either an IMU or an optical tracking ring. 
Subsequent collected data points are then normalized by subtracting this recorded zero-pose value, thereby eliminating relative measurement errors.

Nevertheless, the consistency of the data is maintained as both systems are able to ensure the recording of finger gestures with an error of less than 1 °, and a common calibration standard is strictly applied to ensure the consistency of data annotation.

We used an optical fingerprint scanner to capture touch images at $500$ ppi and $50$ Hz in \emph{RPF} and \emph{CFP}. 
The device supports capturing images of fingers touching the acquisition surface (called plain fingerprint) in single or continuous frames.
In addition, when the finger is continuously rolled on the surface, the device can also stitch the collected results of the entire process into an image, called rolled fingerprint.
Examples of these captures are shown in Figure \ref{fig:db_ex4}.

\subsection{Participants}
Our data collection process of \emph{PRF} is divided into two rounds.
First, a total of $291$ volunteers (243 males, 48 females, aged from $15$ to $50$, $M=20.94, SD=3.07$) were invited to participate in the data collection.
Images of their $3$  frequently used fingers, namely the thumb, index and middle finger of a randomly selected hand (left or right), are recorded.
Subsequently, we recruited other $10$ volunteers (10 males, 0 female, aged from $22$ to $30$, $M=25.10, SD=2.17$) to collect additional data.
Similarly, each participant was asked to capture $3$ frequently used fingers (the thumb, index and middle finger) of both left and right hands.
On the collection process of \emph{CFP}, we recruited $10$ volunteers (10 males, 0 female, aged from $20$ to $26$, $M=25.80, SD=2.18$) to participate in the process.
It should be pointed out that all subjects were strictly ensured not to cross over in the training and testing sets of \emph{RPF} and \emph{CFP}.

\begin{figure}[!t]
	\centering
	\includegraphics[width=.9\linewidth]{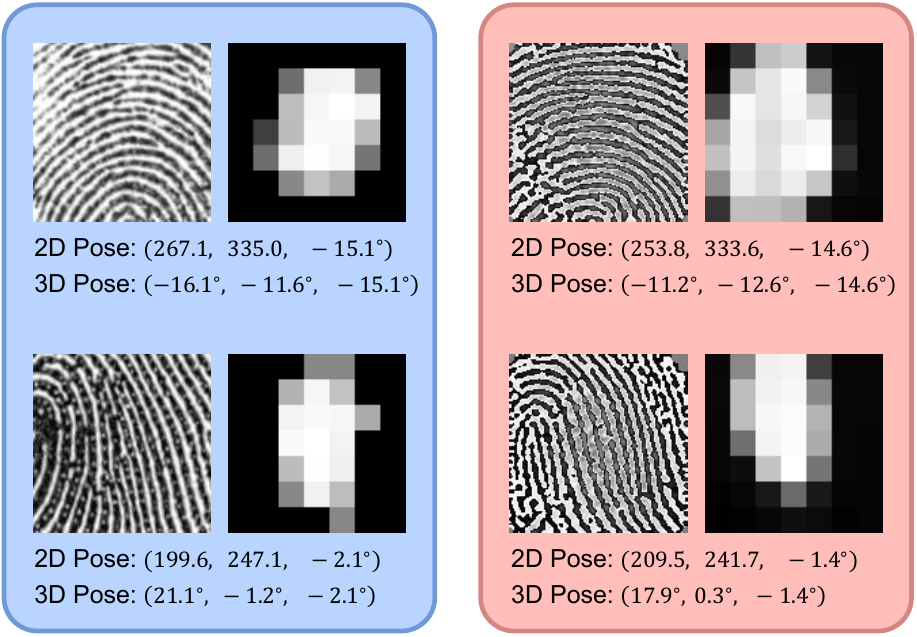}
	\vspace{0mm}
	\caption{Four examples of prepared images and poses. Blue: samples from \emph{RPF}. Red: samples from \emph{CFP}.
    The samples were collected from different volunteers.
    2D and 3D pose is reported as ($x$, $y$, $\theta$) and ($roll$, $pitch$, $yaw$) respectively.}
	\label{fig:generate_ex}
\end{figure}

\subsection{Procedure} 

\begin{figure*}[!t]
	\centering
	\includegraphics[width=.6\linewidth]{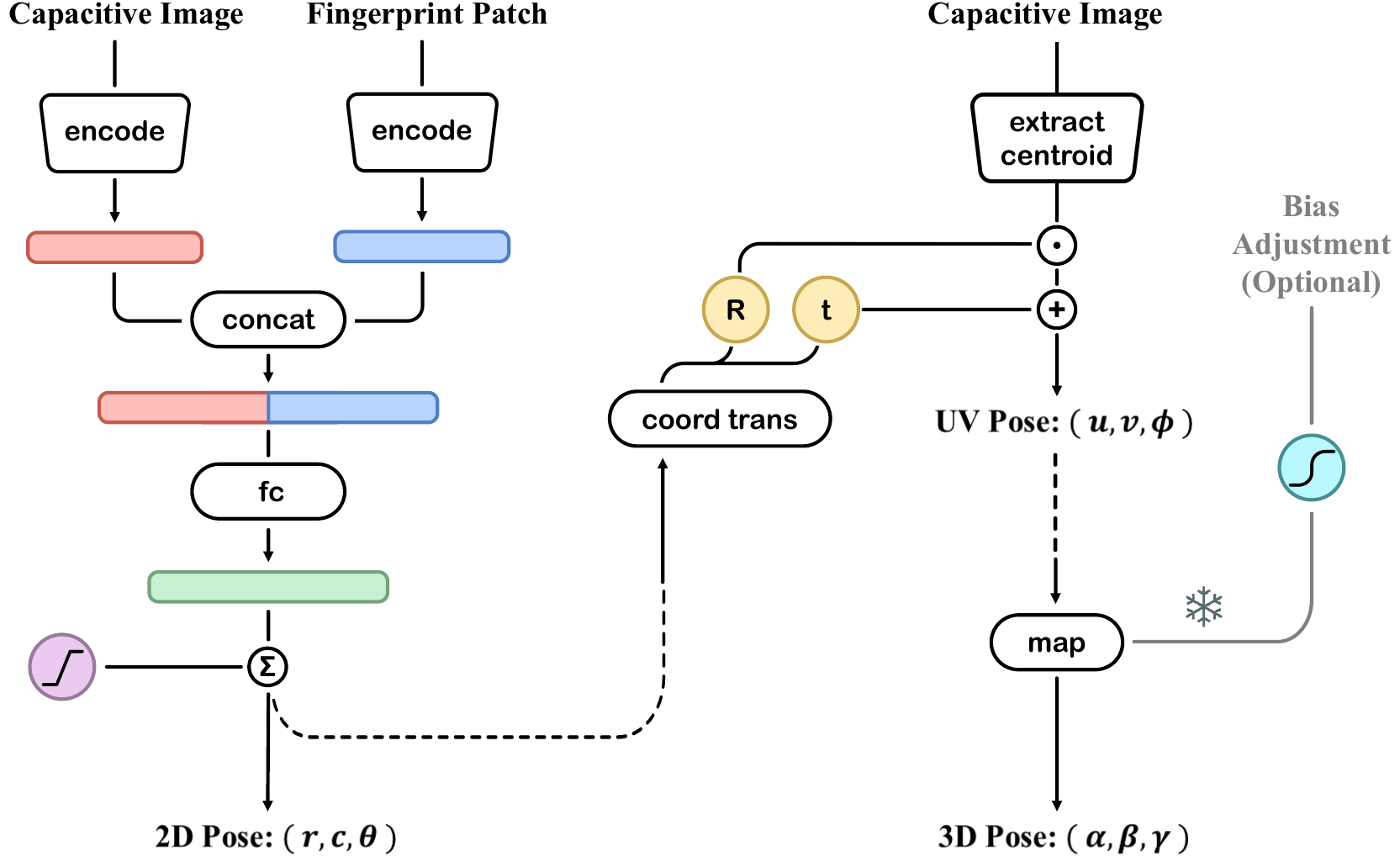}
	\vspace{-2mm}
	\caption{The schematic illustration of our algorithm. 
    The 2D finger pose is initially estimated by our network on the left, then transformed to UV pose via the upper right conversion functions, and finally mapped to 3D pose with the assistance of adjusted freezing parameters. 
    Optionally, users can further enhance the mapping accuracy through a few corrections.}
	\label{fig:network}
\end{figure*}

During the collection process of \emph{PRF}, we utilized the optical fingerprint scanner to capture fingerprint images. 
Participants were instructed to press the device surface in various poses to acquire plain fingerprints.
In addition, they were then guided to roll their fingers from left to right (approximately from $-50^\circ$ to $50^\circ$) to capture rolled fingerprints.
While collecting images, participants were guided to wear the ring-shaped IMU and optical tracking device to acquire 3D pose information, respectively.
Although the equipment used differs, the 3D pose results can be considered unbiased, as they have been meticulously calibrated.
During the collection of \emph{CFP}, volunteers were instructed to perform eight touches at each of the four primary yaw angles \{$0^\circ, 90^\circ, 180^\circ, 270^\circ$\} in any comfortable pose to capture capacitive images and fingerprint patches.
Subsequently, they were directed to collect the corresponding rolled fingerprints.
Finally, they were instructed to capture fingerprint sequences and corresponding 3D poses in a similar manner to \emph{PRF}.

The collected data is preprocessed to obtain a form suitable for training and testing.
Firstly, we employed an existing fingerprint 2D pose estimation method \cite{duan2023estimating} to calibrate the 2D center and rotation angle of the rolled fingerprint automatically.
According to their report, the prediction error in this scenario is very close to the accuracy of manual labeling ($\sim 14\:\mathrm{px}, 5^{\circ}$).
Subsequently, we extracted the minutiae of plain and rolled fingerprints and matched them, which was achieved using existing commercial software \cite{verifinger}.
Let $\boldsymbol{p}$ and $\boldsymbol{q}$ denote the matching minutiae corresponding to paired plain and rolled fingerprints respectively.
The optimal rigid transformation parameters can be obtained by minimizing the reprojection error:
\begin{equation}
	\arg \min_{\theta,t} \sum_{i=1}^{n} \left\|
	\;\boldsymbol{p}_{i}-\left[\begin{array}{cc}
		\cos \theta & -\sin \theta  \\
		\sin \theta & \cos \theta 
	\end{array}\right] 
	\cdot 
	\boldsymbol{q}_{i} - 
	\left[\begin{array}{l}
		t_\mathrm{x} \\
		t_\mathrm{y}
	\end{array}
	\right] \; \right\|_2^2 \;,
	\label{eq:rotation}
\end{equation}
where $n$ is the number of point sets, $\theta$ and $t$ are the rotation angle and translation distance.
Using a plain fingerprint as the reference image, the center and angle of rolled fingerprint can be mapped to obtain the 2D pose of plain fingerprint, as shown in Figure \ref{fig:intro} (b).
Conversely, using the pose-rectified rolled fingerprint as reference image, the foreground center of plain fingerprint (approximately considered as the target contact point) is mapped to obtain the UV pose, as shown in Figure \ref{fig:intro} (c).
Through this paradigm, we successfully linked the ground truth of 2D pose, UV pose, and 3D pose efficiently with the image.
On the other hand, inspired by \cite{streli2021CapContact,maltoni2022handbook,duan2023finger}, we used windowed uniform filtering and interpolation to reduce the fingerprint image to $10$ ppi.
At the same time, we crop the image to $120\times120$ pixels to obtain the fingerprint patch.
The purpose of the above parameter settings is to ensure that the size and resolution of simulated images are consistent with the real dataset \emph{CFP}.
Samples from \emph{CFP} include capacitive images and fingerprint patches, as well as fingerprint frames, corresponding 3D poses and rolled fingerprints.
We employ the same method to obtain the 2D pose and utilize the pose conversion method proposed in Section \ref{subsec:uv-pose-transformation} to obtain the 3D pose.

After the above processing, we have successfully compiled complete samples for both databases \emph{PRF} and \emph{CFP}, encompassing capacitive images, fingerprint patches and 2D/3D poses.
Figure \ref{fig:generate_ex} shows four sets of examples.
The \emph{PRF} comprises 933 fingers from 301 volunteers (10,528 images in total).
Among them, 8,479 images from 744 fingers were utilized for training, 2,049 images of another 189 fingers were employed for testing.
The \emph{CFP} comprises 100 fingers from 10 volunteers (3,200 images in total).
The 3D pose distribution \emph{PRF} and \emph{CFP} is shown in  Figure \ref{fig:db_distribution_RPF} and \ref{fig:db_distribution_CFP}.
Among them, 640 images from 20 fingers were utilized for fine-tuning, another 2,560 images of another 80 fingers were employed for testing.
Specifically, the samples are randomly partitioned and the datasets for fine-tuning and testing are ensured to not contain intersections of the same identity.
More details about training and fine-tuning settings are provided in Section \ref{subsec:Implementation_Details}.

\section{Method}
In this section, we present the proposed BiFingerPose architecture, which enables accurate estimation of 2D and 3D finger pose by leveraging the bimodal input of capacitive images and fingerprint patches.
As shown in Figure \ref{fig:network}, our approach can be divided into three stages:
1) estimating 2D pose through network, 2) transforming 2D pose into UV pose through rigid alignment, 3) mapping UV pose to 3D pose through fitted polynomial function.
Section \ref{subsec:ablation} validates the effectiveness of adopted structural designs and strategies.

\subsection{2D Pose Estimation}\label{subsec:2d-pose-estimation}
Given the tremendous success of deep learning in finger pose estimation \cite{mayer2017estimating,duan2023estimating,Ullerich2023ThumbPitch,he2024trackpose}, a neural network is utilized to perform this task.
Specifically, two encoders with the same structure are used to extract features, and then the two sets of features are concatenated.
ResNext34 \cite{xie2017aggregated} backbone is employed as the feature encoder to fully extract high-level information while avoiding gradient problems.
The extracted information is then flattened and passed through a fully connected layer for further global integration.
Finally, the estimated results are transformed into 2D pose, including the 2D location $(r,c)$ and the angle $\theta$ of the fingerprint center.

A significant difference between our network and previous works is the further evolution of pose information representation \cite{schmitz2021itsy-bits, steuerlein2022conductive}.
Existing deep learning based methods \cite{yin2021joint,he2022estimating,duan2023finger,mayer2017estimating,Ullerich2023ThumbPitch,he2024trackpose} directly regress the numerical values and supervise the mean absolute error or mean square error.
However, while the majority of cases are handled with simplicity and efficacy, there remain potential issues in scenarios involving extensive angle ranges, particularly when angles approach or surpass $180^{\circ}$.
For example, in the case of $\theta = 0^\circ$, both optimization direction of $10^\circ \rightarrow 0^\circ$ and $350^\circ \rightarrow 360^\circ$ will approach the target with the same trend.
The existence of such multiple effective solutions may lead to confusion in network.
To eliminate this ambiguity, we introduce trigonometric function encoding, which is calculated as:
\begin{equation}
	\mathcal{L}_{\mathrm{rot}} = \frac{1}{n} \sum_{i=1}^{n} \left\| \; \hat{cos}_i -cos_i  \; \right\| + \left\| \; \hat{sin}_i-sin_i \; \right\| \;, 
	\label{eq:loss_tan}
\end{equation}
\begin{equation}
	\hat{\theta} = \arctan(\hat{sin}, \hat{cos}) \;,
	\label{eq:loss_tan_pred} 
\end{equation}
where $n$ represents the total number of training samples, $\hat{\theta_i}$ and $\theta_i$ represent the predicted value and ground truth of finger angle,  $\hat{sin}$ and $\hat{cos}$ represent the sine and cosine of target angle.
Moreover, we introduce probability distribution vectors to assist out network in better understanding the adjacency relationships between angles.
The original numerical regression is further evolved into interval classification and weighting, supervised by Cross Entropy ($\mathrm{CE}$) loss.
Let $p$ and $z$ denote the probability and intermediate value of corresponding interval, the final loss function is expressed as:
\begin{equation}
	\mathcal{L}_{\mathrm{rot}} = \mathrm{CE}(\hat{p_\mathrm{cos}}^t,p_\mathrm{cos}^t) + \mathrm{CE}(\hat{p_\mathrm{sin}}^t,p_\mathrm{sin}^t) \;,
	\label{eq:loss_CE} 
\end{equation}
\begin{equation}
	\hat{\theta} = \arctan(\sum_{t=1}^{T}\hat{p_\mathrm{sin}}^t z_\mathrm{sin}^t, \sum_{t=1}^{T}\hat{p_\mathrm{cos}}^t z_\mathrm{cos}^t) \;.
	\label{eq:cal_theta}
\end{equation}
In this paper, we divide the angle interval into $T=120$ equal parts and map them to the numerical value $z$ through corresponding trigonometric functions.
Additionally, Gaussian smoothed labels are used to replace the original one-hot encoding, aiming to help the network better adapt to fuzzy classification boundaries and improve its generalization ability:
\begin{equation}
	p^t(z) = \exp (-\frac{z-z^t}{2\sigma^2}) \;/\; \sum_t p^t(z) \;,
\end{equation}
where $z$ and $z^t$ represent the actual value and the median corresponding to the $t$-th  category, respectively.
Parameter $\sigma$ is empirically f ixed to $2.5$.
Similarly, the center position of 2D finger pose is also supervised through this form, denoted as:
\begin{equation}
	\mathcal{L}_{\mathrm{trans}} = \mathrm{CE}(\hat{p_\mathrm{r}}^t,p_\mathrm{r}^t) + \mathrm{CE}(\hat{p_\mathrm{c}}^t,p_\mathrm{c}^t) \;,
    \label{eq:loss_ce_trans}
\end{equation}
\begin{equation}
	\hat{r} = \sum_{t=1}^{T}\hat{p_\mathrm{r}}^t z_\mathrm{r}^t \;,\;\; \hat{c} =  \sum_{t=1}^{T}\hat{p_\mathrm{c}}^t z_\mathrm{c}^t \;.
    \label{eq:cal_xy}
\end{equation}
The corresponding categories $T$ and variance $\sigma$ are set to $512$ and $3.5$, respectively.

\subsection{UV Pose Transformation}\label{subsec:uv-pose-transformation}

\begin{figure}[!t]
    \centering
    \includegraphics[width=.95\linewidth]{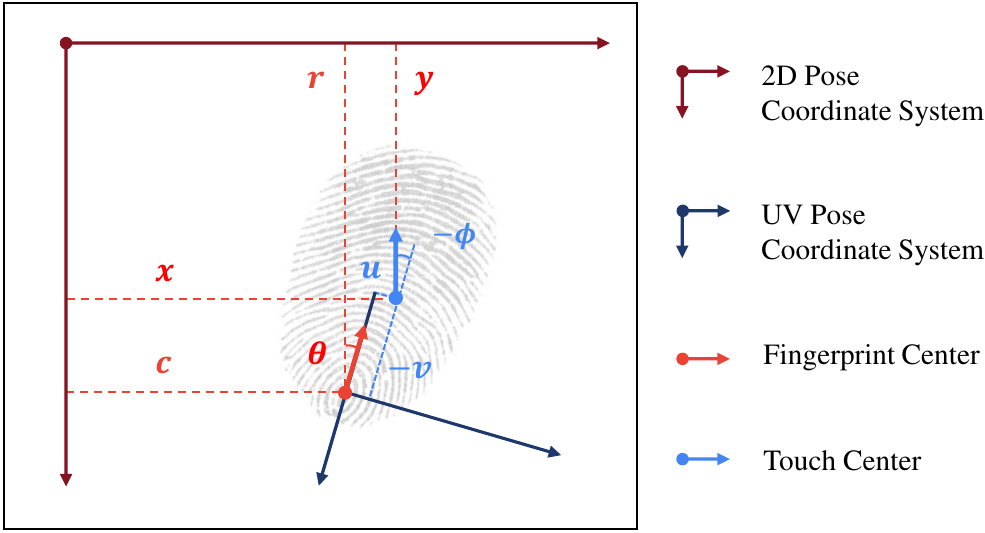} 
    \vspace{-1mm} 
    \caption{
    Schematic diagram of coordinate system transformation from 2D pose to UV pose.
    Information from different coordinate systems is distinguished by color.
    } 
    \label{fig:pose_transformation}
\end{figure}

Drawing upon extensive observations, we have found that the pose of fingers (roll and pitch) tends to be consistent when a specific area is designated as the touch center.
Based on this assumption, we consider fingerprints as textures on finger geometry, and their two-dimensional texture coordinates (referred to as UV coordinate system in this paper) can correspond to a fixed vertex on the three-dimensional object, which is subsequently mapped to 3D angles.
As shown in Figure \ref{fig:intro} (c), the center and positive direction of the fingerprint are used as the coordinate axis center and y-axis positive direction, and the registered pose of the input plain fingerprint is our defined UV pose.

As shown in Figure \ref{fig:pose_transformation}, for a given input sample, we first extract its touch center, which is simply considered as the image center of the fingerprint patch.
Coordinate transformation is subsequently performed to convert the 2D pose (the physiological center and direction of fingerprint in the sensor coordinate system) to the UV pose (the rectified center and direction of touch point in the fingerprint's own reference frame).
Let $(x,y)$ represent the extracted touch center, $(c,r,\theta)$ represent the estimated 2D finger pose, the transformation formula for UV pose is:
\begin{equation}
	\begin{gathered}
		R=\left[\begin{array}{cc}
			\cos \theta & \sin \theta  \\
			-\sin \theta & \cos \theta 
		\end{array}\right] \;,\;\;
		t= - R \cdot \left[\begin{array}{l}
			c \\
			r
		\end{array}
		\right] \;,\\
		\left[\begin{array}{l}
			u \\
			v
		\end{array}
		\right] = R \cdot 
		\left[\begin{array}{l}
			x \\
			y
		\end{array}
		\right] + t \;, \;\; \phi = -\theta \;.
	\end{gathered}
\end{equation}
where $(u,v)$ and $\phi$ is the 2D position and angle of touch center in the UV coordinate system.
This conversion step corresponds to the double-headed arrow between Figure \ref{fig:intro} (b) and (c). Since this is a rigid transformation, no additional approximation errors are introduced.

\subsection{3D Pose Mapping} \label{subsec:3d-pose-mapping}
In this subsection, we describe how to establish a precise mapping relationship between UV pose and 3D pose.
Polynomial functions of one and two variables are selected as the target curve families to be fitted, which are defined as:
\begin{equation}
	\begin{gathered}
		\alpha(u,v) = \sum_{i=1}^{k} \sum_{j=0}^{i}a_{ij} u^j v^{i-j}+ b \;, \\ 
		\beta(u,v) = \sum_{i=1}^{k} \sum_{j=0}^{i}a_{ij} u^j v^{i-j}+ b \;, \\
		\gamma = \phi + b\;,
	\end{gathered}
	\label{eq:map}
\end{equation}
where $\alpha$, $\beta$, $\gamma$ correspond to roll, pitch and yaw angle of 3D pose respectively, $(u,v,\phi)$ is the UV pose, $k$ is the highest power of corresponding polynomial, while $a$ and $b$ are the parameters to be fitted.
Any feasible optimization algorithm can be employed for curve fitting, such as the nonlinear least squares method utilized in this paper.
Additionally, we propose two mechanisms for establishing mappings to meet the needs of different scenarios:
\begin{enumerate}
	\item \textbf{Global Optimization}: All training samples are used to calculate the global optimal parameters and directly deploy them in real products without registration, which can be regarded as a baseline.
	\item \textbf{Global Optimization with Adaption}: On the basis of global optimization parameters ${a_{ij}}$ and $b$, requesting users to register only a few times ($1\sim4$ touches) to adjust the bias (as shown on the right side of Figure \ref{fig:network}).
\end{enumerate}
After thorough comparison and discussion, we ultimately decide to use the quartic function to fit the mapping relationship, while using mechanism (2) as determined parameter adjustment scheme.

\section{Evaluation}

\subsection{Implementation Details} \label{subsec:Implementation_Details}
In our finalized network, the ResNext-34 backbone \cite{xie2017aggregated} serves as the encoder for the branches of the two modalities.
On the one hand, a $7\times7$ capacitve image (cropped according to the interactive area) is input.
After passing through the encoder (without downsampling, with intermediate channels set to ${32,64,128,256,512}$), a feature map of size (512,7,7) is obtained. 
Global average pooling (GAP) is then applied, and the result is flattened into a 512-dimensional vector.
On the other hand, a $120\times120$ fingerprint patch is fed into the other encoder, which operates with downsampling and has intermediate channels configured as ${64,128,256,512, 1024}$.
This produces a feature map of size (1024,3,3). 
Similarly, GAP is applied and the result is flattened into a 1024-dimensional vector.
The two corresponding vectors are concatenated (resulting in a 1536-dimensional vector) and then linearly transformed to 512 dimensions through a fully connected layer. 
The first half and the other half of the class probability information are passed through two softmax layers and used to compute the horizontal and vertical position of 2D finger pose, respectively, following the calculation in Equation \ref{eq:cal_xy}.
Additionally, the concatenated vector is passed through another fully connected layer and two softmax layers to produce a 120-dimensional probability vector, where the first half and the second half correspond to the sine and cosine values, respectively.
The angle of the 2D finger pose is then calculated using Equation \ref{eq:cal_theta}.
The 2D pose predicted by our network is then mapped to the 3D pose using the non-learning method introduced in Sections \ref{subsec:uv-pose-transformation} and \ref{subsec:3d-pose-mapping}.

The training process is conducted on the training set of \emph{RPF} with an initial learning rate of 1e-3 (reduced to 1e-6), a cosine annealing scheduler, the default AdamW optimizer, and a batch size of 128, continuing until convergence (approximately 80 epochs).
During the evaluation, we partitioned the \emph{RPF} test set into three intervals based on the yaw angle ($|\gamma|\leq 45^\circ/90^\circ/135^\circ/180^\circ$) to fully assess the performance of the algorithm in different scenarios. 
Consequently, during training, we applied random angle rotations as data augmentation, while ensuring they were constrained within the specified angle ranges. 
In other words, the same model incorporates multiple weights, each tailored for evaluation within the corresponding angle range.
When evaluating in \emph{CFP}, we initialize the model with the pre-trained weights from \emph{RPF} ($|\gamma|\leq 180^\circ$) and perform a few-shot fine-tuning.
The learning rate is gradually reduced from 1e-4 to 1e-6, and other settings are the same as the training stage.
For detailed information on the dataset construction and partitioning please refer to Section \ref{sec:data-collection}.

\subsection{Baseline Methods} \label{subsec:baseline}
We employ representative 2D/3D pose estimation methods that are free from dense registration and applicable to capacitive images or fingerprint patches as baselines for comparison.
First, we employ the 42 hand-crafted features defined in \cite{xiao2015estimating} to estimate finger pose, which uses capacitive image as input. 
To fully leverage the potential of large-scale data, we replace the original gaussian regression process with a 5-layer multi-layer perceptron (MLP) with intermediate channel sizes of ${64, 128, 256, 512, 3}$.
In the following we refer to it as \emph{MLP}.
Inspired by \cite{mayer2017estimating,Ullerich2023ThumbPitch,he2024trackpose}, we also use a convolutional neural network (CNN) to predict finger pose from capacitive images.
Given the small size of their original model, a direct comparison might be unfair.
Therefore, we replace their network with ResNeXt-34 \cite{xie2017aggregated} and refer to it as \emph{CNN}.
It should be noted that these original implementations \cite{xiao2015estimating,mayer2017estimating,Ullerich2023ThumbPitch,he2024trackpose} do not predict the roll angle, and we explored addressing this limitation by adding relevant prediction parameters.
On the other hand, we utilize networks from \cite{yin2021joint} and \cite{duan2023estimating}, which excel in the fingerprint modality, to estimate finger pose from fingerprint patches. 
As the size of these models is similar to ours, we directly adopt the same architecture proposed by them, labeled \emph{JointNet} and \emph{GridNet}, respectively.
To the best of our knowledge, the above works are currently the most prominent and state-of-the-art methods in this field.
The training and testing configurations for all the aforementioned models are consistent with our method. 

\subsection{Evaluation Metrics}
We employ mean absolute error (MAE), root mean square error (RMSE), and standard deviation (SD) to evaluate the performance of 2D and 3D pose estimation, respectively. 
Given that the yaw angle of the 3D pose and the angle of the 2D pose are equivalent, the final evaluation encompasses five parameters:  the roll, pitch, and yaw angles of the 3D pose (see Figure \ref{fig:intro}a), and the horizontal (u) and vertical (v) positions of the UV pose (see Figure \ref{fig:intro}c).
It should be mentioned that we employ the horizontal and vertical coordinates in the UV pose instead of the 2D pose estimation. 
This transformation is lossless.
More importantly, the position in the UV pose is based on the finger's own coordinates, which is more intuitive.
Reporting both poses as metrics is necessary because some algorithms are originally designed for 2D pose \cite{yin2021joint,duan2023estimating} and others for 3D pose \cite{xiao2015estimating, mayer2017estimating,Ullerich2023ThumbPitch,he2024trackpose}.

\begin{table*}[!t]
    
    \belowrulesep=-0.2pt
	\aboverulesep=-0.2pt
	\caption{Quantitative results for 2D (u, v and yaw) and 3D (yaw, pitch and roll) finger pose estimation in \emph{RPF}.
Position (u and v) and angular (yaw, pitch and roll) errors are reported in pixels and degrees respectively.}
	\label{tab:RPF}
	\vspace{-0.1cm}
    \setlength\tabcolsep{4pt}
    \begin{center}
    
    \begin{tabular}{c|c|ccc|ccc|ccc|ccc|ccc}
    \toprule
    \multicolumn{1}{c|}{\multirow{2}{*}{\makecell[c]{\textbf{\scriptsize{Yaw Angle}}\\\textbf{\scriptsize{Interval}}}}}
    &\multicolumn{1}{c|}{\multirow{2}{*}{\makecell[c]{\textbf{\scriptsize{Method}}}}}
    & \multicolumn{3}{c|}{\textbf{\scriptsize{u}}}
    & \multicolumn{3}{c|}{\textbf{\scriptsize{v}}}
    & \multicolumn{3}{c|}{\textbf{\scriptsize{yaw}}}
    & \multicolumn{3}{c|}{\textbf{\scriptsize{pitch}}}
    & \multicolumn{3}{c}{\textbf{\scriptsize{roll}}}\\
    \cmidrule(lr){3-5}\cmidrule(lr){6-8}\cmidrule(lr){9-11}\cmidrule(lr){12-14}\cmidrule(lr){15-17}
    \multicolumn{1}{c|}{}
    & \multicolumn{1}{c|}{}
    & \multicolumn{1}{c}{\textbf{\scriptsize{MAE}}}
    & \multicolumn{1}{c}{\textbf{\scriptsize{RMSE}}}
    & \multicolumn{1}{c|}{\textbf{\scriptsize{SD}}}
    & \multicolumn{1}{c}{\textbf{\scriptsize{MAE}}}
    & \multicolumn{1}{c}{\textbf{\scriptsize{RMSE}}}
    & \multicolumn{1}{c|}{\textbf{\scriptsize{SD}}}
    & \multicolumn{1}{c}{\textbf{\scriptsize{MAE}}}
    & \multicolumn{1}{c}{\textbf{\scriptsize{RMSE}}}
    & \multicolumn{1}{c|}{\textbf{\scriptsize{SD}}}
    & \multicolumn{1}{c}{\textbf{\scriptsize{MAE}}}
    & \multicolumn{1}{c}{\textbf{\scriptsize{RMSE}}}
    & \multicolumn{1}{c|}{\textbf{\scriptsize{SD}}}
    & \multicolumn{1}{c}{\textbf{\scriptsize{MAE}}}
    & \multicolumn{1}{c}{\textbf{\scriptsize{RMSE}}}
    & \multicolumn{1}{c}{\textbf{\scriptsize{SD}}}\\
    \midrule
    \multicolumn{1}{c|}{\multirow{5}{*}{\makecell[c]{{\tiny{{$[-45^\circ,45^\circ]$}}}}}} 
    & \scriptsize{MLP \cite{xiao2015estimating}} & \scriptsize{50.1} & \scriptsize{63.5} & \scriptsize{39.1} & \scriptsize{46.9} & \scriptsize{57.7} & \scriptsize{33.7} & \scriptsize{12.4} & \scriptsize{15.8} & \scriptsize{9.7} & \scriptsize{8.5} & \scriptsize{10.5} & \scriptsize{6.2} & \scriptsize{18.5} & \scriptsize{23.5} & \scriptsize{14.5} \\
    {} & \scriptsize{CNN \cite{mayer2017estimating,Ullerich2023ThumbPitch,he2024trackpose}} & \scriptsize{36.9} & \scriptsize{46.4} & \scriptsize{28.2} & \scriptsize{43.9} & \scriptsize{56.4} & \scriptsize{35.3} & \scriptsize{7.2} & \scriptsize{9.2} & \scriptsize{5.7} & \scriptsize{8.0} & \scriptsize{10.3} & \scriptsize{6.5} & \scriptsize{13.7} & \scriptsize{17.2} & \scriptsize{10.4} \\
    {} & \scriptsize{JointNet \cite{yin2021joint}} & \scriptsize{24.3} & \scriptsize{32.7} & \scriptsize{21.8} & \scriptsize{30.0} & \scriptsize{42.4} & \scriptsize{29.9} & \scriptsize{16.6} & \scriptsize{20.9} & \scriptsize{12.6} & \scriptsize{5.4} & \scriptsize{7.6} & \scriptsize{5.4} & \scriptsize{9.0} & \scriptsize{12.1} & \scriptsize{8.1} \\
    {} & \scriptsize{GridNet \cite{duan2023estimating}} & \scriptsize{20.2} & \scriptsize{27.4} & \scriptsize{18.5} & \scriptsize{29.2} & \scriptsize{37.5} & \scriptsize{23.6} & \scriptsize{11.6} & \scriptsize{14.3} & \scriptsize{8.4} & \scriptsize{5.3} & \scriptsize{6.8} & \scriptsize{4.3} & \scriptsize{7.5} & \scriptsize{10.1} & \scriptsize{6.9} \\
    {} & \scriptsize{ours} & \textbf{\scriptsize{12.5}} & \textbf{\scriptsize{16.8}} & \textbf{\scriptsize{11.3}} & \textbf{\scriptsize{13.4}} & \textbf{\scriptsize{18.9}} & \textbf{\scriptsize{13.3}} & \textbf{\scriptsize{5.7}} & \textbf{\scriptsize{7.2}} & \textbf{\scriptsize{4.4}} & \textbf{\scriptsize{2.4}} & \textbf{\scriptsize{3.4}} & \textbf{\scriptsize{2.4}} & \textbf{\scriptsize{4.6}} & \textbf{\scriptsize{6.2}} & \textbf{\scriptsize{4.2}} \\
    \hline

    \multicolumn{1}{c|}{\multirow{5}{*}{\makecell[c]{{\tiny{{$[-90^\circ,90^\circ]$}}}}}} 
    & \scriptsize{MLP \cite{xiao2015estimating}} & \scriptsize{51.0} & \scriptsize{64.5} & \scriptsize{39.5} & \scriptsize{41.3} & \scriptsize{52.1} & \scriptsize{31.7} & \scriptsize{38.9} & \scriptsize{47.8} & \scriptsize{27.7} & \scriptsize{7.5} & \scriptsize{9.5} & \scriptsize{5.8} & \scriptsize{18.9} & \scriptsize{23.9} & \scriptsize{14.6} \\
    {} & \scriptsize{CNN \cite{mayer2017estimating,Ullerich2023ThumbPitch,he2024trackpose}} & \scriptsize{39.6} & \scriptsize{51.0} & \scriptsize{32.0} & \scriptsize{42.9} & \scriptsize{55.2} & \scriptsize{34.7} & \scriptsize{12.8} & \scriptsize{18.8} & \scriptsize{13.7} & \scriptsize{7.8} & \scriptsize{10.0} & \scriptsize{6.3} & \scriptsize{14.7} & \scriptsize{18.9} & \scriptsize{11.9} \\
    {} & \scriptsize{JointNet \cite{yin2021joint}} & \scriptsize{26.7} & \scriptsize{35.8} & \scriptsize{23.8} & \scriptsize{30.1} & \scriptsize{41.4} & \scriptsize{28.4} & \scriptsize{17.7} & \scriptsize{22.8} & \scriptsize{14.4} & \scriptsize{5.4} & \scriptsize{7.5} & \scriptsize{5.2} & \scriptsize{9.9} & \scriptsize{13.2} & \scriptsize{8.8} \\
    {} & \scriptsize{GridNet \cite{duan2023estimating}} & \scriptsize{35.1} & \scriptsize{46.4} & \scriptsize{30.3} & \scriptsize{32.7} & \scriptsize{44.2} & \scriptsize{29.7} & \scriptsize{18.3} & \scriptsize{23.8} & \scriptsize{15.2} & \scriptsize{5.9} & \scriptsize{8.0} & \scriptsize{5.4} & \scriptsize{13.0} & \scriptsize{17.2} & \scriptsize{11.2} \\
    {} & \scriptsize{ours} & \textbf{\scriptsize{12.7}} & \textbf{\scriptsize{17.4}} & \textbf{\scriptsize{11.8}} & \textbf{\scriptsize{14.6}} & \textbf{\scriptsize{20.5}} & \textbf{\scriptsize{14.3}} & \textbf{\scriptsize{6.1}} & \textbf{\scriptsize{8.4}} & \textbf{\scriptsize{5.8}} & \textbf{\scriptsize{2.7}} & \textbf{\scriptsize{3.7}} & \textbf{\scriptsize{2.6}} & \textbf{\scriptsize{4.7}} & \textbf{\scriptsize{6.4}} & \textbf{\scriptsize{4.4}} \\
    \hline

    \multicolumn{1}{c|}{\multirow{5}{*}{\makecell[c]{{\tiny{{$[-135^\circ,135^\circ]$}}}}}} 
    & \scriptsize{MLP \cite{xiao2015estimating}} & \scriptsize{51.7} & \scriptsize{65.0} & \scriptsize{39.4} & \scriptsize{44.4} & \scriptsize{57.0} & \scriptsize{35.7} & \scriptsize{65.5} & \scriptsize{75.7} & \scriptsize{38.0} & \scriptsize{8.1} & \scriptsize{10.3} & \scriptsize{6.4} & \scriptsize{19.1} & \scriptsize{24.0} & \scriptsize{14.6} \\
    {} & \scriptsize{CNN \cite{mayer2017estimating,Ullerich2023ThumbPitch,he2024trackpose}} & \scriptsize{52.4} & \scriptsize{66.1} & \scriptsize{40.3} & \scriptsize{50.0} & \scriptsize{62.2} & \scriptsize{36.9} & \scriptsize{58.5} & \scriptsize{73.6} & \scriptsize{44.6} & \scriptsize{9.0} & \scriptsize{11.2} & \scriptsize{6.6} & \scriptsize{19.4} & \scriptsize{24.4} & \scriptsize{14.9} \\
    {} & \scriptsize{JointNet \cite{yin2021joint}} & \scriptsize{32.0} & \scriptsize{43.6} & \scriptsize{29.6} & \scriptsize{32.1} & \scriptsize{43.3} & \scriptsize{29.0} & \scriptsize{24.9} & \scriptsize{32.7} & \scriptsize{21.2} & \scriptsize{5.8} & \scriptsize{7.8} & \scriptsize{5.3} & \scriptsize{11.8} & \scriptsize{16.1} & \scriptsize{11.0} \\
    {} & \scriptsize{GridNet \cite{duan2023estimating}} & \scriptsize{36.6} & \scriptsize{47.9} & \scriptsize{30.8} & \scriptsize{42.6} & \scriptsize{53.8} & \scriptsize{32.8} & \scriptsize{35.7} & \scriptsize{44.9} & \scriptsize{27.3} & \scriptsize{7.7} & \scriptsize{9.7} & \scriptsize{5.9} & \scriptsize{13.6} & \scriptsize{17.7} & \scriptsize{11.4} \\
    {} & \scriptsize{ours} & \textbf{\scriptsize{13.4}} & \textbf{\scriptsize{18.6}} & \textbf{\scriptsize{12.9}} & \textbf{\scriptsize{15.2}} & \textbf{\scriptsize{21.5}} & \textbf{\scriptsize{15.3}} & \textbf{\scriptsize{6.2}} & \textbf{\scriptsize{8.6}} & \textbf{\scriptsize{6.0}} & \textbf{\scriptsize{2.8}} & \textbf{\scriptsize{3.9}} & \textbf{\scriptsize{2.8}} & \textbf{\scriptsize{5.0}} & \textbf{\scriptsize{6.9}} & \textbf{\scriptsize{4.8}} \\
    \hline

    \multicolumn{1}{c|}{\multirow{5}{*}{\makecell[c]{{\tiny{{$[-180^\circ,180^\circ]$}}}}}} 
    & \scriptsize{MLP \cite{xiao2015estimating}} & \scriptsize{54.5} & \scriptsize{68.8} & \scriptsize{41.9} & \scriptsize{51.2} & \scriptsize{66.3} & \scriptsize{42.1} & \scriptsize{91.1} & \scriptsize{103.2} & \scriptsize{48.5} & \scriptsize{9.3} & \scriptsize{12.0} & \scriptsize{7.6} & \scriptsize{20.2} & \scriptsize{25.4} & \scriptsize{15.5} \\
    {} & \scriptsize{CNN \cite{mayer2017estimating,Ullerich2023ThumbPitch,he2024trackpose}} & \scriptsize{57.9} & \scriptsize{73.2} & \scriptsize{44.9} & \scriptsize{56.0} & \scriptsize{69.3} & \scriptsize{40.7} & \scriptsize{70.4} & \scriptsize{83.5} & \scriptsize{44.9} & \scriptsize{10.2} & \scriptsize{12.6} & \scriptsize{7.4} & \scriptsize{21.4} & \scriptsize{27.1} & \scriptsize{16.6} \\
    {} & \scriptsize{JointNet \cite{yin2021joint}} & \scriptsize{39.2} & \scriptsize{51.9} & \scriptsize{34.0} & \scriptsize{35.7} & \scriptsize{48.4} & \scriptsize{32.6} & \scriptsize{30.8} & \scriptsize{40.7} & \scriptsize{26.6} & \scriptsize{6.5} & \scriptsize{8.8} & \scriptsize{5.9} & \scriptsize{14.5} & \scriptsize{19.2} & \scriptsize{12.6} \\
    {} & \scriptsize{GridNet \cite{duan2023estimating}} & \scriptsize{54.5} & \scriptsize{67.2} & \scriptsize{39.3} & \scriptsize{49.7} & \scriptsize{64.4} & \scriptsize{40.8} & \scriptsize{64.5} & \scriptsize{79.6} & \scriptsize{46.8} & \scriptsize{9.0} & \scriptsize{11.7} & \scriptsize{7.5} & \scriptsize{20.1} & \scriptsize{24.8} & \scriptsize{14.6} \\
    {} & \scriptsize{ours} & \textbf{\scriptsize{14.5}} & \textbf{\scriptsize{19.8}} & \textbf{\scriptsize{13.6}} & \textbf{\scriptsize{15.7}} & \textbf{\scriptsize{21.9}} & \textbf{\scriptsize{15.3}} & \textbf{\scriptsize{6.6}} & \textbf{\scriptsize{8.8}} & \textbf{\scriptsize{5.9}} & \textbf{\scriptsize{2.8}} & \textbf{\scriptsize{4.0}} & \textbf{\scriptsize{2.8}} & \textbf{\scriptsize{5.4}} & \textbf{\scriptsize{7.3}} & \textbf{\scriptsize{5.0}}\\

    \bottomrule
    \end{tabular}
    \end{center}
\end{table*}

\begin{table*}[!t]
    \belowrulesep=-0.2pt
	\aboverulesep=-0.2pt
	\caption{Quantitative results for 2D (u, v and yaw) and 3D (yaw, pitch and roll) finger pose estimation in \emph{CFP}.
Position (u and v) and angular (yaw, pitch and roll) errors are reported in pixels and degrees respectively.}
	\label{tab:CFP}
	\vspace{-0.1cm}
    \setlength\tabcolsep{4pt}
    \begin{center}
    \begin{tabular}{c|ccc|ccc|ccc|ccc|ccc}
    \toprule
    \multicolumn{1}{c|}{\multirow{2}{*}{\makecell[c]{\textbf{\scriptsize{Method}}}}}
    & \multicolumn{3}{c|}{\textbf{\scriptsize{u}}}
    & \multicolumn{3}{c|}{\textbf{\scriptsize{v}}}
    & \multicolumn{3}{c|}{\textbf{\scriptsize{yaw}}}
    & \multicolumn{3}{c|}{\textbf{\scriptsize{pitch}}}
    & \multicolumn{3}{c}{\textbf{\scriptsize{roll}}}\\
    \cmidrule(lr){2-4}\cmidrule(lr){5-7}\cmidrule(lr){8-10}\cmidrule(lr){11-13}\cmidrule(lr){14-16}
    \multicolumn{1}{c|}{}
    & \multicolumn{1}{c}{\textbf{\scriptsize{MAE}}}
    & \multicolumn{1}{c}{\textbf{\scriptsize{RMSE}}}
    & \multicolumn{1}{c|}{\textbf{\scriptsize{SD}}}
    & \multicolumn{1}{c}{\textbf{\scriptsize{MAE}}}
    & \multicolumn{1}{c}{\textbf{\scriptsize{RMSE}}}
    & \multicolumn{1}{c|}{\textbf{\scriptsize{SD}}}
    & \multicolumn{1}{c}{\textbf{\scriptsize{MAE}}}
    & \multicolumn{1}{c}{\textbf{\scriptsize{RMSE}}}
    & \multicolumn{1}{c|}{\textbf{\scriptsize{SD}}}
    & \multicolumn{1}{c}{\textbf{\scriptsize{MAE}}}
    & \multicolumn{1}{c}{\textbf{\scriptsize{RMSE}}}
    & \multicolumn{1}{c|}{\textbf{\scriptsize{SD}}}
    & \multicolumn{1}{c}{\textbf{\scriptsize{MAE}}}
    & \multicolumn{1}{c}{\textbf{\scriptsize{RMSE}}}
    & \multicolumn{1}{c}{\textbf{\scriptsize{SD}}}\\
    \midrule
    \scriptsize{MLP \cite{xiao2015estimating}} & \scriptsize{37.2} & \scriptsize{48.2} & \scriptsize{30.6} & \scriptsize{69.0} & \scriptsize{83.8} & \scriptsize{47.7} & \scriptsize{75.9} & \scriptsize{91.9} & \scriptsize{51.9} & \scriptsize{12.6} & \scriptsize{15.3} & \scriptsize{8.7} & \scriptsize{13.8} & \scriptsize{17.8} & \scriptsize{11.3} \\
    \scriptsize{CNN \cite{mayer2017estimating,Ullerich2023ThumbPitch,he2024trackpose}} & \scriptsize{47.8} & \scriptsize{59.4} & \scriptsize{35.3} & \scriptsize{73.8} & \scriptsize{91.8} & \scriptsize{54.6} & \scriptsize{68.1} & \scriptsize{78.7} & \scriptsize{39.4} & \scriptsize{13.4} & \scriptsize{16.7} & \scriptsize{10.0} & \scriptsize{17.7} & \scriptsize{22.0} & \scriptsize{13.1} \\
    \scriptsize{JointNet \cite{yin2021joint}} & \scriptsize{37.2} & \scriptsize{50.0} & \scriptsize{33.4} & \scriptsize{44.1} & \scriptsize{60.9} & \scriptsize{42.0} & \scriptsize{35.2} & \scriptsize{47.2} & \scriptsize{31.5} & \scriptsize{8.0} & \scriptsize{11.0} & \scriptsize{7.6} & \scriptsize{13.8} & \scriptsize{18.5} & \scriptsize{12.3} \\
    \scriptsize{GridNet \cite{duan2023estimating}} & \scriptsize{42.4} & \scriptsize{54.1} & \scriptsize{33.6} & \scriptsize{69.6} & \scriptsize{86.5} & \scriptsize{51.4} & \scriptsize{63.2} & \scriptsize{78.0} & \scriptsize{45.7} & \scriptsize{12.6} & \scriptsize{15.7} & \scriptsize{9.4} & \scriptsize{15.7} & \scriptsize{20.0} & \scriptsize{12.4} \\
    \scriptsize{ours} & \textbf{\scriptsize{18.6}} & \textbf{\scriptsize{28.4}} & \textbf{\scriptsize{21.5}} & \textbf{\scriptsize{19.4}} & \textbf{\scriptsize{32.9}} & \textbf{\scriptsize{26.6}} & \textbf{\scriptsize{11.9}} & \textbf{\scriptsize{17.9}} & \textbf{\scriptsize{13.4}} & \textbf{\scriptsize{3.5}} & \textbf{\scriptsize{5.9}} & \textbf{\scriptsize{4.8}} & \textbf{\scriptsize{6.9}} & \textbf{\scriptsize{10.5}} & \textbf{\scriptsize{8.0}} \\
 
    \bottomrule
    \end{tabular}
    \end{center}
\end{table*}

\subsection{Comparison Results}

\begin{figure}[!t]
    \centering
    \includegraphics[width=.85\linewidth]{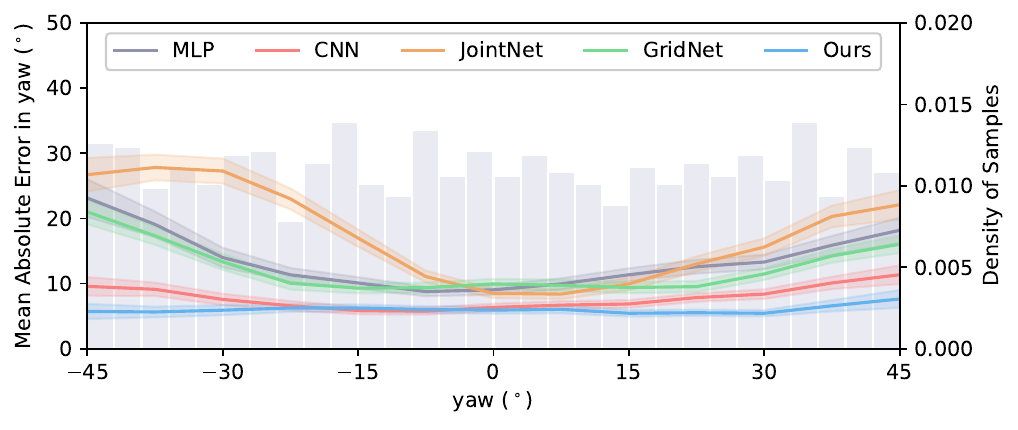} 
    \vspace{-1mm} 
    \includegraphics[width=.85\linewidth]{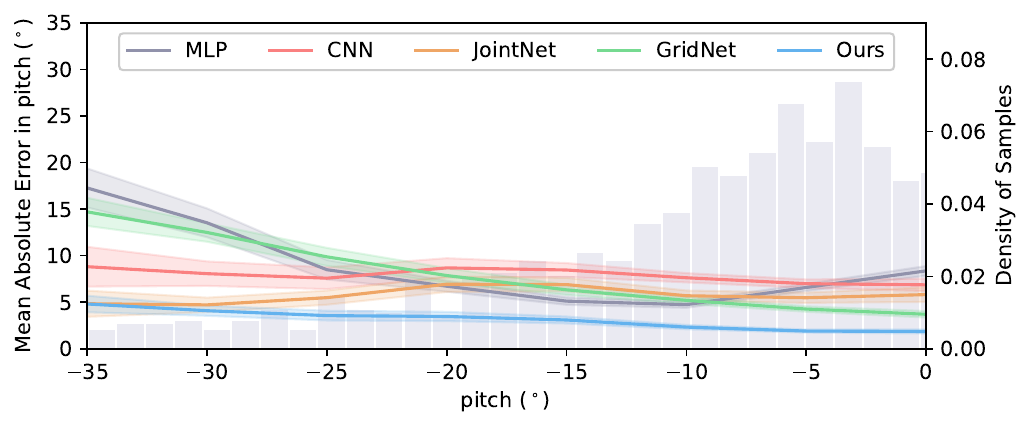} 
    \vspace{-1mm} 
    \includegraphics[width=.85\linewidth]{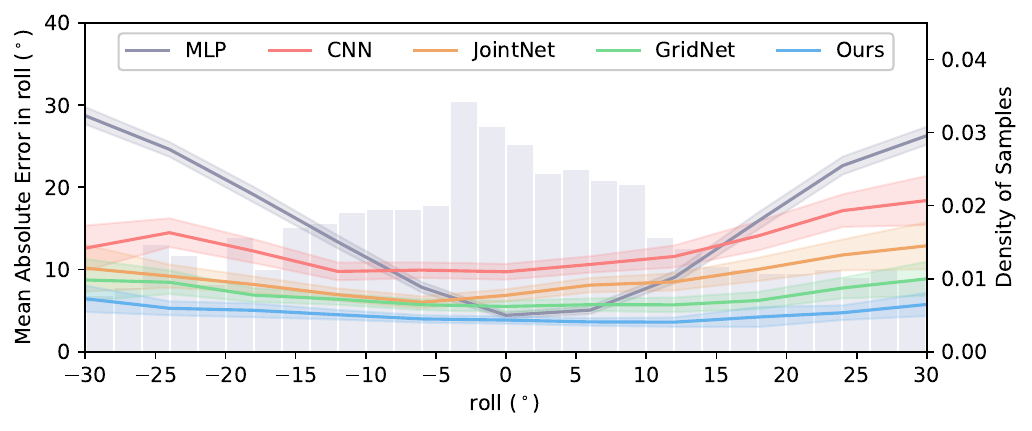}
    \vspace{-2mm} 
    \caption{Error distribution of 3D finger pose in the \emph{RPF} test set (small-angle interactive scene). 
    95\% confidence interval (CI) is shown as color bar.} 
    \label{fig:ex_distribution}
\end{figure}

Table \ref{tab:RPF}  and \ref{tab:CFP} provide a comparative evaluation of algorithm performance over varying yaw angle ranges in the \emph{RPF} / \emph{CFP} test set.
The results indicate that capacitive image based algorithms  (\emph{MLP}, \emph{CNN}) achieve superior performance at small yaw angles, whereas the fingerprint patch based approaches (\emph{JointNet}, \emph{GridNet}) exhibit distinct advantages in 2D pose positioning and roll/pitch angle estimation.
Furthermore, the prediction errors of these unimodal methods increase significantly as the yaw angle range expands.
Our proposed BiFingerPose demonstrates superior performance compared to these state-of-the-art methods, achieving significant improvements in both 2D and 3D pose estimation accuracy.
Remarkably, the proposed method retains outstanding stability and precision even under extremely large yaw angles, substantially outperforming other single-modal approaches.
Across all metrics presented in Tables \ref{tab:RPF} and \ref{tab:CFP}, BiFingerPose consistently achieves a \textbf{performance improvement exceeding 21\%} when compared to the suboptimal baseline method. 
Notably, the experimental results also confirm that our method successfully estimates the roll angle with high accuracy, which is not supported by the baseline methods.
Figure \ref{fig:ex_distribution} further illustrates the error distribution across different algorithms under small-angle conditions, demonstrating the consistent advantages of our method.
This substantial enhancement not only validates the effectiveness of BiFingerPose but also highlights its significant potential for practical implementation across diverse real-world applications.

\subsection{Ablation Study} \label{subsec:ablation}
Above all, we conduct experiments on the test set of \emph{RPF} with the yaw angle range of $[-180^\circ,180^\circ]$ to examine how representation and supervision strategies influence the performance of 2D finger pose estimation networks.
In terms of rotation angle in 2D pose, the average error of our network (in fingerprint modality) is $22.5^\circ$ when using the original regression supervision. 
This error decreases to $12.1^\circ$ after incorporating triangulation and is further reduced to $3.1^\circ$ with the implementation of our proposed full loss function (see Equation \ref{eq:loss_CE}).
On the other hand, the localization error of 2D pose decreases from $34.3 \:\text{px}$ in the regression form to $31.4\:\text{px}$ (as Equation \ref{eq:loss_ce_trans}) when using the Euclidean distance.
It can be seen that the evolution of angle towards trigonometric form significantly improves its prediction accuracy.
Furthermore, the introduction of probability distribution vectors demonstrates notable positive effects on both angle and position estimation.

Next, we validate the effectiveness of modality fusion. 
In the pose estimation network depicted in Figure \ref{fig:network}, when only the capacitive image branch is utilized, the angle estimation and localization error of 2D finger pose are $68.6^\circ$ and $74.3\:\text{px}$, respectively. 
When only the fingerprint patch branch is employed, the errors are $14.1^\circ$ and $25.4\:\text{px}$. 
However, when the bimodal is used, errors are reduced to $6.55^\circ$ and $23.9\:\text{px}$.
This result strongly demonstrates the effectiveness of our proposed bimodal approach.

Finally, we assess the potential errors that may arise during the conversion between 2D and 3D pose.
As outlined in Section \ref{subsec:uv-pose-transformation}, the conversion from 2D pose to UV pose is a non-parametric linear transformation that is error-free.
During the conversion from UV pose to 3D pose, mechanism (1) from Section \ref{subsec:3d-pose-mapping} results in average conversion errors of $13.2^\circ$ and $7.3^\circ$ for roll and pitch angle, respectively. 
When employing  mechanism (2), the errors are effectively reduced to $3.2^\circ$ and $1.9^\circ$.
Therefore, we strongly recommend the users implement simple adjustments to minimize errors during the pose transition phase (if necessary).

\section{User Study} \label{sec:user-study}

\subsection{Baseline Methods}
For comprehensive evaluation, we compare our bimodal based BiFingerPose against two state-of-the-art single modal based approaches: (1) capacitive image based 
\emph{CNN} inspired by \cite{mayer2017estimating,Ullerich2023ThumbPitch,he2024trackpose}, and (2) fingerprint patch based \emph{GridNet} reimplemented from \cite{duan2023estimating}. 
Both baseline methods demonstrate superior performance on the corresponding single modality in small yaw angle scenarios (see Table \ref{tab:RPF}). 
It is worth noting that other relatively weaker baseline protocols mentioned in Section \ref{subsec:baseline} were excluded from this experiment to avoid overburdening subjects with excessive testing.

\subsection{Tasks}

\begin{figure*}[!t]
	\centering
	\includegraphics[width=.9\linewidth]{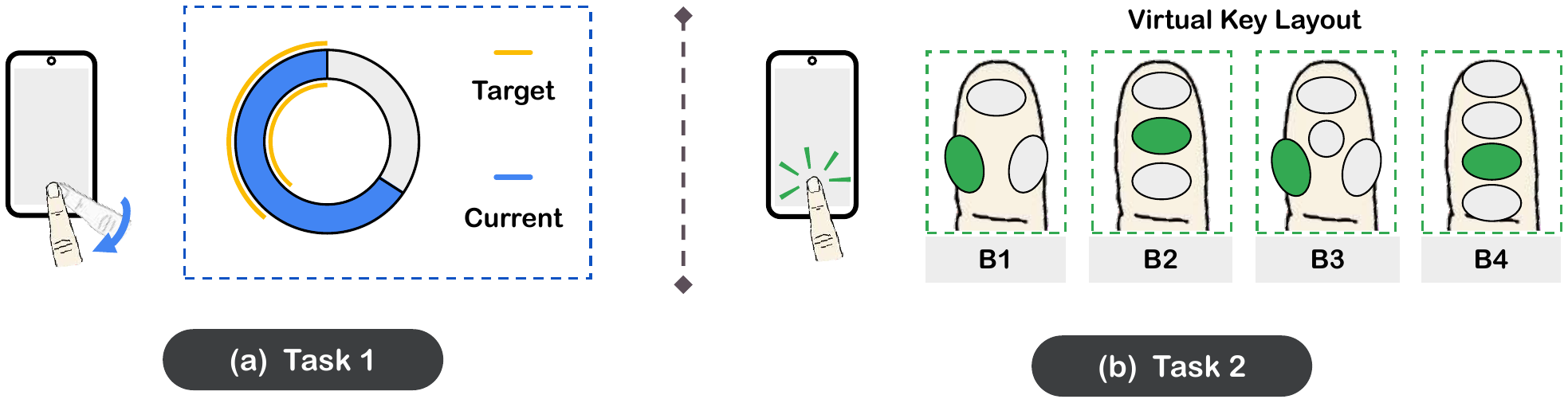}
	\vspace{-1mm}
	\caption{Schematic diagrams of user study scenarios: (a) users rotate their fingers to adjust the progress bar until they reach the target, (b) users touch the phone with their finger at a designated position based on a certain prompt.
	Contents in the dashed box represent the guiding information displayed for each task.}
	\label{fig:ex_user:user}
\end{figure*}

Two tasks operated on mobile phones are implemented to evaluate feasibility and performance in terms of completion time and accuracy:
\begin{itemize}
	\item Progress bar adjustment based on finger rotation to check continuous control performance.
	\item Command input based on finger position to reflect discrete interaction performance.
\end{itemize}
In addition, questionnaire survey and interview are conducted to evaluate the user experience and subjective feelings of participants.

Figure \ref{fig:ex_user:user} shows the corresponding schematic diagram.
While integrating diverse pose parameters facilitates a range of tasks, maintaining consistent accuracy and stability remains paramount, forming the cornerstone of our Task 1 evaluation.
On the other hand, by recognizing different fingers through fingerprints, up to $8 \times 4 = 32$ functional areas can be mapped in layout B3 (thumb excluded), which can be further extended to the common QWERTY keyboard ($26$ keys).
Inspired by the 9-key layout, which first selects the main group and then makes specific choices, we can even provide a richer vocabulary for discrete interactions by introducing coordinated gestures such as sliding.
Anyway, such tasks typically use the input accuracy as a general metric,  which is measured in our Task 2.
We believe that these two fundamental and widely applicable tasks can represent typical usage patterns encountered on most portable electronic devices, thus reflecting the general performance in diverse scenarios.

\subsection{Apparatus}
The experiment was conducted in a quiet office.
The same smartphone described in Section \ref{subsec:data_collection-apparatus} was used to capture capacitive images and fingerprint patches simultaneously.
Guidance information is displayed on a computer, as shown in Figure \ref{fig:ex_user:user}.
Participants were informed that they could hold the phone with any comfortable gesture and complete these designated tasks without other restrictions

\subsection{Participants}
We recruited $12$ volunteers ($9$ males, $3$ females, aged from 15 to 40, $M=25.8$, $SD=5.7$) through social media to take part in our user study. 
Nine of them are right-handed.
All participants were advised to use their dominant hand for operation and underwent approximately $20$ minutes of training and adaptation before experiments.
It should be noted that the identities of all participants in this section have no overlap with the participants used as training sets.

\subsection{Procedure}
\subsubsection{Task 1: Progress bar adjustment based on finger rotation}
In this experiment, participants were asked to control the progress bar by rotating their fingers in 3D space, with the objective of achieving a predetermined target.
As shown in Figure \ref{fig:ex_user:user}a, the target progress bar is randomly set between $0$\% and $100$\% with a fixed random seed, while the user controlled bar is initialized at $50$\%.
The relationship between angle and value is linear, and a clockwise rotation corresponds to decrement in progress.
This task will be considered as successfully achieved within an error of $\pm\:2\%$ and maintained for $2$ seconds.
Considering universality and acceptability, we use yaw angle as a representative, as previous SOTA methods \cite{duan2023estimating,mayer2017estimating,Ullerich2023ThumbPitch,he2024trackpose} do not support predicting the other one or two angles with sufficient accuracy.
In addition, we also separately measured the average completion time when utilizing roll and pitch based on BiFingerPose, to compare the relative manipulation performance of individual 3D finger angles.
Each participant performed $10$ times using the above three pose estimation methods under the same random seed.

\begin{figure*}[!t]
	\centering
	\subfloat[\small{Task 1}]{\includegraphics[height=4.5cm]{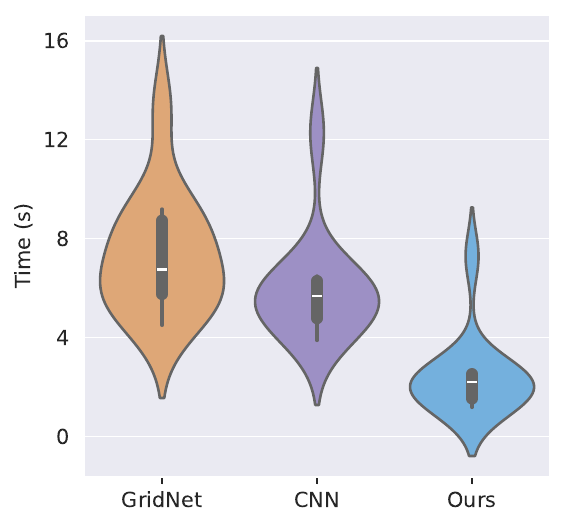}%
        \label{fig:ex_user:rotation}
	\vspace{-1mm}}
	\hfil
	\subfloat[\small{Task 2}]{\includegraphics[height=4.5cm]{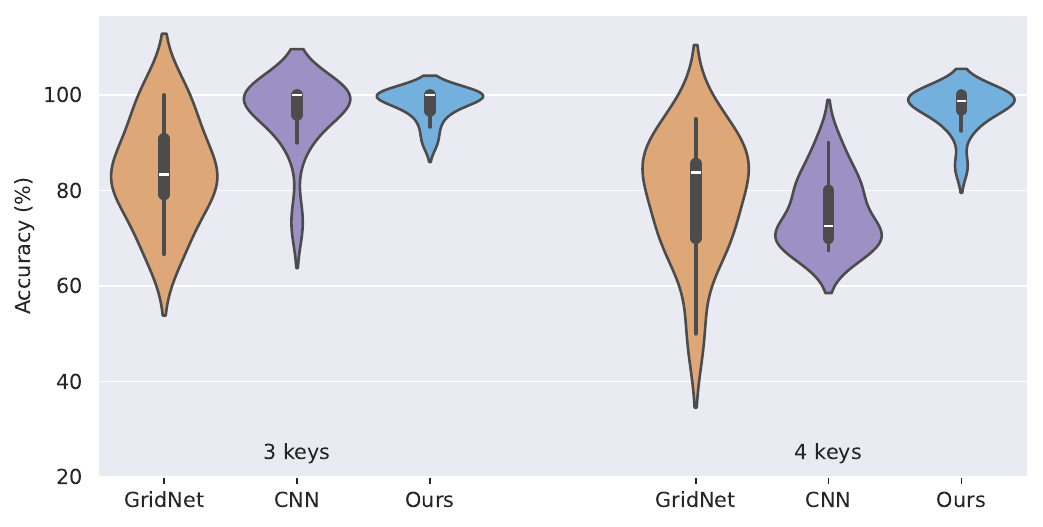}%
	\label{fig:ex_user:position}
        \vspace{-1mm}}
        \label{fig:ex_user}
	\caption{Quantitative evaluation of user study on two tasks: (a) Completion times for continuous control in Task 1. (b) Accuracy of for discrete interaction in Task 2.}
\end{figure*}

\begin{figure*}[!t]
	\centering
	\subfloat[\small{Task 1}]{\includegraphics[width=0.48\textwidth]                     {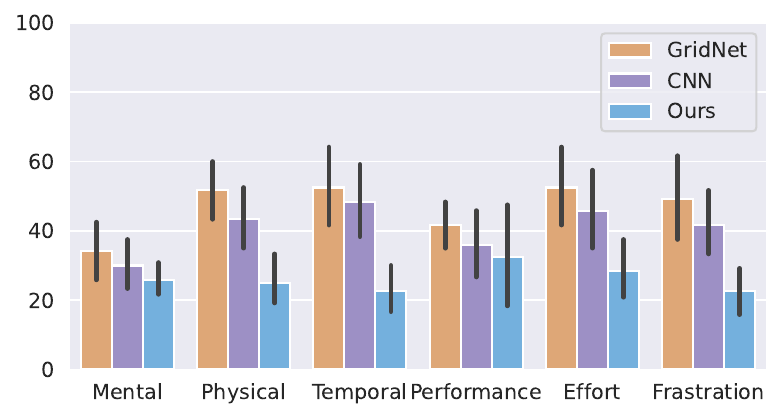}%
	\vspace{-1mm}}
	\hfil
	\subfloat[\small{Task 2}]{\includegraphics[width=0.48\textwidth]{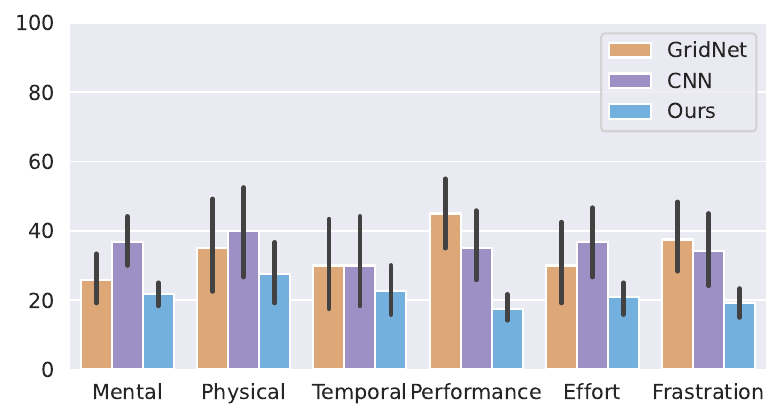}%
        \vspace{-1mm}}

	\caption{NASA-TLX questionnaire results (0-100).
		Lower score indicates lower mental / physical / temporal demand, higher performance, lower effort / frustration. 
		Error bars: 95\% CI.}
	\label{fig:ex_user:measurement}
\end{figure*}

\subsubsection{Task 2: Command input based on finger position}
This task aims to evaluate the performance of 2D finger pose estimation in discrete interactions, utilizing representative command scenario (in the form of virtual keyboards) as a benchmark.
Participants are required to input designated keys (finger positions) based on guidance information.
Due to limited finger space, we divide it into $3$ or $4$ key positions, as shown in Figure \ref{fig:ex_user:user}b.
Each key has $10$ tests, and the order of appearance between all keys is random, with a fixed random seed for each person.
It should be noted that by disregarding the yaw angle (to ensure compatibility with diverse grip postures), \emph{CNN} only measures one signal of vertical position parameter, as its horizontal position prediction performance is relatively insufficient (see Table \ref{tab:RPF}). 
As a result, it can only use virtual keyboards in the form of Figure \ref{fig:ex_user:user} B2 and B4.
In contrast, the other two methods are capable of measuring two dimensions (2D position) in addition to the yaw angle, allowing for a more dispersed key position of Figure \ref{fig:ex_user:user} B1 and B3.
The layout design intuitively emphasizes the importance of introducing roll angle to expand the interactive space.

\subsection{Comparison Results}

Shapiro-Wilk test is used to examine the normality of data and questionnaires across all tasks.
Results indicate that only the subjective rating follow a normal distribution.
Therefore, we use the bar chart with $95$ \% confidence interval (CI) to display the feedback of user experience, while presenting violin plots and tables to reflect other performance metrics.

\subsubsection{Task 1: Evaluation of continuous control performance}

Figure \ref{fig:ex_user:rotation} shows the distribution of completion time for this continuous control task.
ANOVA are used to verify the differences in task completion time among the three methods.
According to the order of \emph{GridNet}-\emph{Ours}, \emph{CNN}-\emph{Ours}, the statistical results are as follows: $(F=34.31, p<.0001)$, and $(F=20.88, p<.05)$.
The results indicate significant differences among our method and baseline solutions.
The average time consumption of each method is $7.4$ s for \emph{GridNet}, $6.0$ s for \emph{CNN}, and $2.4$ s for the proposed BiFingerPose.
In other words, the completion efficiency of our approach is $\boldsymbol{2.5}$ \textbf{times faster} than the SOTA method for 3D pose estimation under the same operation time, while about $\boldsymbol{3.1}$ \textbf{times faster} than the SOTA method for 2D pose estimation.
The average completion times for pitch and roll angle of BiFingerPose are $3.0$ s and $2.2$ s, respectively.
The performance comparison among three angles emphasizes the significant advantage  of roll angle in interaction efficiency, which have not yet been shown on mobile phones but can be effectively utilized through our bimodal approach.
Additionally, we observed that participants spent a lot of time fine-tuning the progress bar while completing projects based on \emph{GridNet} and \emph{CNN}, which proves the necessity of sufficient precision for fine interaction.

\subsubsection{Task 2: Evaluation of discrete interaction performance}
The performance of virtual keyboard input based on finger pose mapping is shown in Figure \ref{fig:ex_user:position}.
Here we only focus on the relative accuracy differences between three pose estimation methods, and leave further exploration with typing solutions to interested researchers.
In the $3$-key scenario, the average accuracy of \emph{GridNet}, \emph{CNN}, and our BiFingerPose is $84.2 \%$, $96.1 \%$ and $98.1 \%$, respectively.
The ANOVA results show significant evidence of a difference in error rates between the our method and \emph{GridNet} ($F=18.9, p<.05$), while not significant compared to \emph{CNN} ($F=0.62, p=.44$).
In the $4$-key scenario, the performance of all methods shows a certain decline, with accuracy of $78.8 \%$, $75.4 \%$ and $97.1 \%$.
However, our method still maintains a high level of precision and consistently leads the way ($\boldsymbol{23\%}$ \textbf{higher accuracy} than suboptimal method).
ANOVA test shows that our method has significant differences from both baselines ($F=22.14, P<.05$ and $F=75.5, p<.0001$, respectively).
These results indicate that the proposed method outperforms previous advanced methods in discrete control.
Specifically, \emph{CNN} experiences a pronounced deterioration in performance as the number of keys expands, which makes sense as the excessively compact layout are more likely to cause confusion in nearby positions.
To some extent, this also highlights the benefits that a more comprehensive input dimension (roll angle in this paper) can bring to the expansion of interaction space.

\subsubsection{Subjective evaluation}

For convenience and intuitive feedback, participants were asked to filled out the NASA-TLX  \cite{hart1988development} questionnaire to assess the perceived workload under different pose estimation methods, as reported in Figure \ref{fig:ex_user:measurement}.
To quantify the user’s overall subjective feeling, we utilize the inverse of the average score in questionnaires as the overall indicator (ranging from 0 to 100, where higher values indicate better performance).
The \emph{GridNet} and \emph{CNN} achieved scores of 56.1 and 58.6, respectively, while our approach was rated 74.5, a \textbf{27\% improvement} over the optimal baseline.
Results indicate that participants overwhelmingly appreciate the engaging interactive experience afforded by our BiFingerPose, with \emph{CNN} being the subsequent refer.

As part of the interview procedure, we encouraged participants to freely express their views and perspectives.
Among them, $9$ participants (P1, P3-P5,P8-P12) stated that the finger pose based interaction method was "\emph{easy to grasp and intuitive to use}" .
P3 said, "\emph{using finger pose for interaction gives me a different experience from buttons and touch. When rotating my fingers, I can naturally control the angle even without observing the feedback on screen.}"
The potential application of virtual keyboard in Task 2 has also aroused users' interest, "\emph{by specifying the finger area and combining simple actions such as sliding, perhaps I can call out various applications on the smart watch through individual gestures instead of multiple operations,}" commented P1.
Most of the participants (P1, P3, P4, P6-P9, P11) believe that the introduction of roll angles is beneficial to improving operational efficiency and interactive experience.
Compared to \emph{GridNet} and \emph{CNN}, almost all users (P1, P3-P12) indicated that the user experience of BiFingerPose is "the most comfortable and enjoyable", while expressing a clear preference for this model.
P2 and P7 suggested that achieving a higher refresh rate can further improve the smoothness of operation, which is about $15$ Hz due to certain limitations of the prototype device.
Finally, all users expressed affirmation and expectation for the prospect of introducing finger pose interaction, especially for lightweight portable devices such as smartwatches, wristbands, rings, etc.

\begin{figure}[!t]
	\centering
	\includegraphics[width=.95\linewidth]{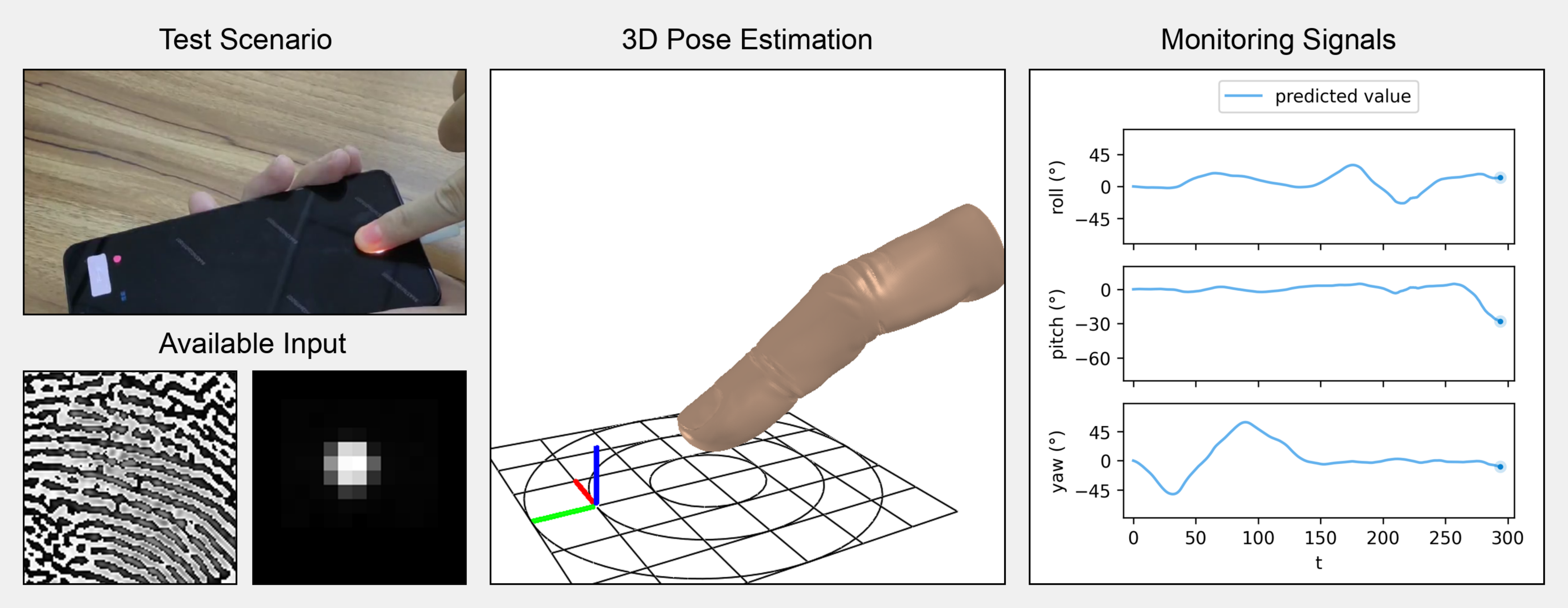}
	\vspace{-3mm}
	\caption{Example of one frame in our video demonstration, which is the visualization result of our pose estimation method applied on mobile phone with under screen optical fingerprint sensor.}
	\label{fig:ex_visual}
\end{figure}

\section{Application Scenarios} \label{sec:application-scenarios}

Firstly, the pose estimation process of our proposed BiFingerPose is visualized \textbf{in our supplementary video}.
Two video clips of 2D (in UV pose format for appropriateness) and 3D pose estimation are presented, such as Figure \ref{fig:ex_visual}, to intuitively demonstrate its overall performance in finger pose prediction.
When visualizing 2D pose, we draw a collection box and center symbol based on the relative position relationship, with the pose rectified rolled fingerprint as center (also the background).
On the other hand, a 3D finger model was used to visualize the 3D pose.

\begin{figure*}[!t]
	\centering
	\includegraphics[width=.6\linewidth]{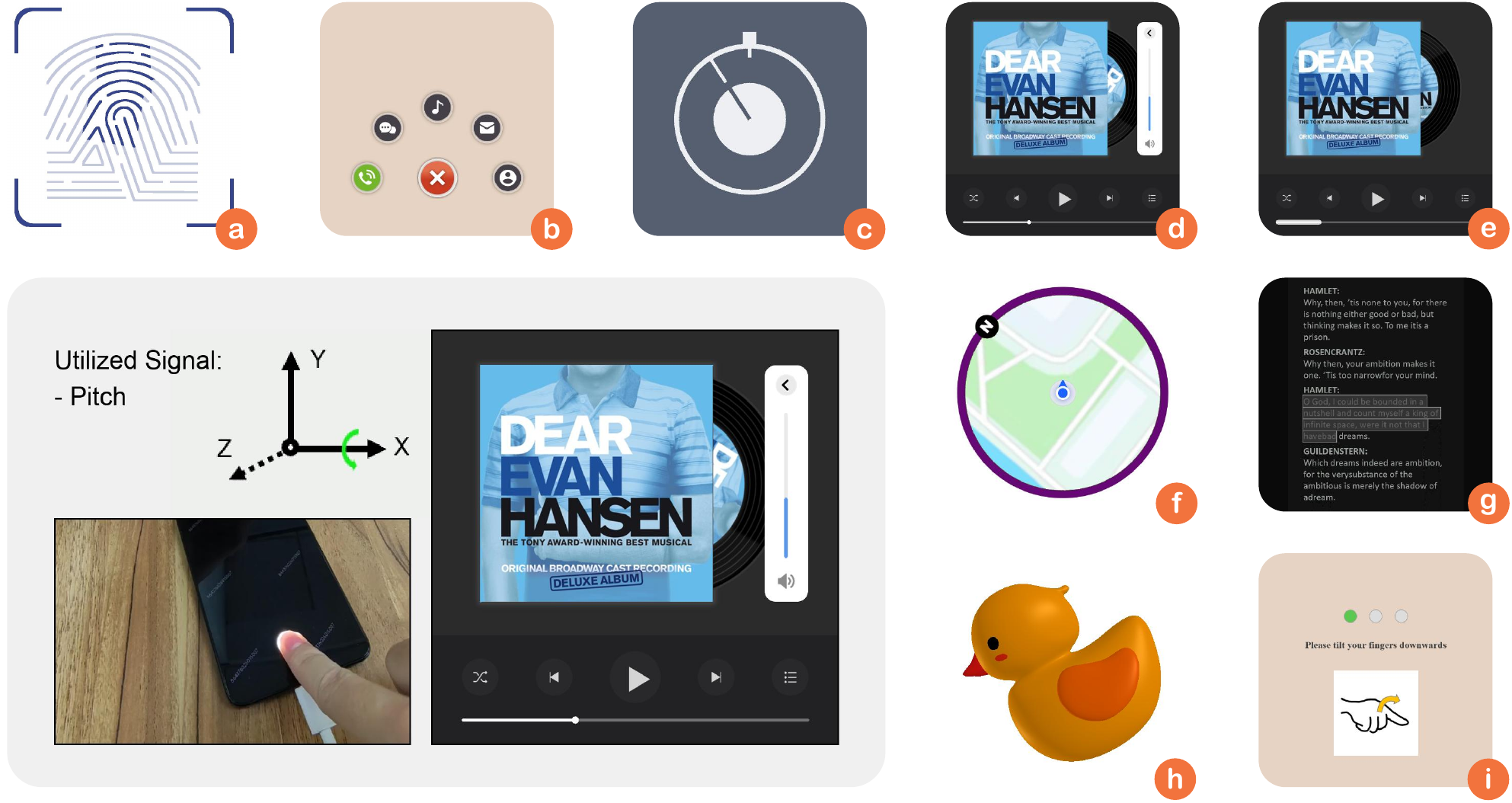}
	\vspace{-3mm}
	\caption{Examples of potential application scenarios The complete interface is displayed in the bottom left corner, including prompts for utilized signals, user operation, and interactive effects.}
	\label{fig:application}
\end{figure*}

To demonstrate the expansive value of finger pose in real life applications, we explore and discuss its potential functionality in multiple deployment scenarios, and provide corresponding prototypes in video form as intuitive examples (\textbf{represented in the supplementary video}).
An overview of these demos is shown in Figure \ref{fig:application}.
According to the conversion signals applied, applications are roughly classified into three categories, namely position of 2D pose, single angle of 3D pose, and combination of multiple pose information.
We believe this is a beneficial supplement to \cite{vogelsang2021design} that allows the audience to better understand its practical value in various aspects and providing possible inspiration for future interface and deployment designs.

\subsection{Position of 2D Pose}
To the best of our knowledge, existing fingerprint registration programs in mobile phones only requires users to provide a specified number of fingerprints, without precise guidance \cite{maltoni2022handbook}.
The addition of location information can assist users in knowing which finger positions have not been covered yet, as shown in Figure \ref{fig:application} (a), thereby avoiding duplication or omission of fingerprint areas.
On the other hand, by mapping finger positioning to virtual keyboards and integrating it with existing simple gestures, a variety of selection wheels can be readily summoned and operated, enabling quick access to a large number of applications and even efficient typing in limited interaction spaces.
Figure \ref{fig:application} (b) illustrates a corresponding example, where user activates the desired application group by pressing a designated finger region, and then slides to choose the "phone" function.
This allows users to easily access multiple function selection wheels through a single touch of different contact regions, whereas existing smartwatches usually require several swipes until the desired function group interface is reached.

\subsection{Single Angle of 3D Pose}
The design for single angle interaction is advised to align with the target audience’s intuition of physical sensation \cite{vogelsang2021design}.
For instance, in the case of the timer depicted in Figure \ref{fig:application} (c), counterclockwise rotation is closer to the general cognition and more conducive to starting without comprehension barriers, thus showing greater advantages than sliding or button based adjustments.
In addition, Figures \ref{fig:application} (d) and (e) show the control process of volume and progress through pitch and roll angles, respectively.
Owing to the extensive practice in daily activities, users can effortlessly forge a seamless and precise correlation between finger angles and their corresponding linear controllers.

\subsection{Combination of Multiple Pose Information}
The combination of multiple pose information can assist in handling more complex and sophisticated tasks.
In the example of Figure \ref{fig:application} (f), user can control the rotation and scaling of map through yaw and pitch angle within a very limited interaction area.
Besides, the combination of pitch and roll is used to select text across multiple rows and columns in Figure \ref{fig:application} (g), while avoiding occlusion.
Moreover, Figure \ref{fig:application} (h) illustrates the direct mapping of 3D pose to control the orientation of rubber duck, which can be regarded as a typical representative of 3D manipulation.
Finally, a security application is provided in Figure \ref{fig:application} (i), which performs live detection by randomly specifying and checking verification actions to avoid malicious attacks and spoofing.
We emphasize that compared to the most common interaction implementation of virtual keyboards with multi-touch, solutions described here can be efficiently implemented in confined spaces, which has particular value on mobile devices with limited size or full-size interactions that need to avoid occlusion as much as possible (such as smart watch).

\subsection{Subjective Evaluation}

\begin{figure}[!t]
	\centering
	\includegraphics[width=.9\linewidth]{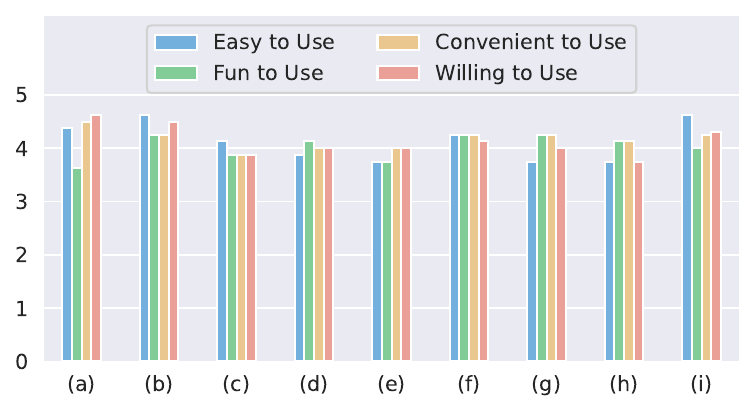}
	\vspace{-3mm}
	\caption{Subjective ratings of finger pose in different applications (0-5). 
		Higher score indicate easier to use, more fun to use, more convenient to use, and more willing to use.
		The labels correspond to the applications shown in Figure \ref{fig:application}.}
	\label{fig:application_measurement}
\end{figure}

The same participants of Section \ref{sec:user-study} were invited to evaluate the above applications.
Figure \ref{fig:application_measurement} present the subjective ratings of these application scenarios in four dimensions.
Statistical results show that participants generally enjoy the interactive experience enabled by BiFingerPose, with an average will to use of $4.1$.
Among them, participants show special attention to security related applications, with (a)  and (i) ranking first ($4.6$) and third ($4.3$) respectively.
Independent Action based interactions, such as (a), (b) and (i), are considered easier to use compared with most continuous interactions (corresponding average scores are $4.6$ and $3.9$, respectively).
In addition, the combination of pose information (application (f), (g) and (i)) is believed to provide greater convenience, as it demonstrates satisfactory applicability to various complex scenarios.
An interesting observation is that the feedback on directly manipulating 3D objects is relatively flat. 
One possible explanation is that there is a lack of relevant applications in existing mobile devices, making it difficult for evaluators to associate an intuitive usage scenario through imagination.

\section{Discussion and Limitations}

Compared to existing solutions that achieve similar interaction functions but rely on additional devices \cite{streli2022TapType,zhang2022typeanywhere,liang2021DualRing,liu2023printype}, our method is deployable purely through software updates on any touch system equipped with an under-screen fingerprint sensor, thereby quickly and effectively expanding its input dimensions and enhancing the user experience.
Specifically, BiFingerPose can be integrated on any touch device equipped with an under-screen fingerprint sensor, provided that the system possesses sufficient computational resources to run a 10M parameter deep learning model.
Moreover, our method diverges from traditional registration and matching techniques \cite{holz2010generalized, duan2022estimating, liu2023printype} by directly generating reliable predictions through the robust generalization of deep learning. This not only streamlines the process but also circumvents potential privacy risks associated with handling fingerprint data.
Additionally, when the device is occasionally used by others (such as family members and friends), a registration-free experience is highly desirable.
On the other hand, compared to previous works on finger pose estimation \cite{yin2021joint,duan2023estimating,xiao2015estimating,mayer2017estimating,Ullerich2023ThumbPitch,he2024trackpose}, our approach exhibits outstanding stability and precision.
Especially, our proposed BiFingerPose performs well even within a $360^\circ$ yaw angle input range.
We believe that these are very important considerations when users considering whether to choose a certain interaction solution.

The presented work still leaves some scope for further improvement.
Firstly, we observed small fluctuations in BiFingerPose in a few tests.
We believe that this can be alleviated by adding appropriate rule constraints in targeted scenarios.
For example, in applications of finger pitch, short-term and high-frequency action recognition (lift or fall) can be utilized instead of original numerical form of estimation, which may increase the resistance to uncontrolled small fluctuations to a certain extent.
In addition, introducing temporal information \cite{he2024trackpose} or contrastive learning \cite{dai2020rankpose} may also be helpful for continuous prediction.
Secondly, the scene of sensors can be further explored. 
In this paper, we conducted experimentson a smartphone equipped with a under screen fingerprint sensor of specific size.
The performance at some other resolutions and shapes, such as slender rectangles captured by some phones' side fingerprint sensors, still needs to be carefully measured for appropriate usages.
Additionally, our current system speed is relatively low (15 Hz due to some I/O limitations in current device). 
User experience can be further improved through hardware deployment optimization to enhance system processing speed.
Furthermore, we believe that in scenarios with small input space, such as smartwatches/bracelets/rings (where glasses can be used as separate display terminals), the advantages of finger pose interaction will be more prominent. 
Engineering issues of the algorithm deployment on these devices need further exploration.
Besides, this method can also be deployed on mobile phones equipped with large area underscreen fingerprint sensors \cite{qualcomm}, providing a broader interactive space.
Moreover, for enhanced generalization to complex and practical scenarios, addressing challenges posed by occlusion and multi-user (specifically multi-finger) environments is imperative. 
Potential solutions encompass utilizing image restoration methodologies and incorporating identity recognition based on fingerprint patches.
Finally, while our dataset covers a wide range of identities and poses, more exploration is needed to address the challenges of finger modality variations (e.g., dry, wet, aging conditions) and to collect and validate larger-scale datasets. 
In the future, we will focus on these aspects to perfect the user experience of our proposed BiFingerPose.

\section{Conclusion}
We present BiFingerPose, a bimodal finger pose estimation framework for touch devices.
We explored a new bimodal of capacitive image and fingerprint patch that can be directly applied on existing mobile devices and highlight its remarkable advantages.
Meanwhile, triangulated probability distribution vector is introduced to replace the regression form output in previous networks, significantly enhancing both prediction accuracy and stability.
Moreover, we demonstrate that basic polynomial functions are adequate to reliably map 2D pose into the 3D angle domain.
By adopting this conversion paradigm, researchers can readily utilize existing finger pose estimation algorithms, initially designed for a specific definition, to facilitate interactive tasks that may be better suited to alternative pose definitions.
Extensive experiments affirm that BiFingerPose surpasses the previous state-of-the-art finger pose estimation algorithms, especially at large angles.
In addition, by introducing the fingerprint patch modality, we also make roll angle possible in interactive applications, which is currently not supported by capacitive modality alone.
A 12-person user study further showcases the superiority of our solution in terms of interaction efficiency and subjective experience.
To round out our presentation, we engage in a detailed discussion and provided prototype videos across a range of application scenarios, fully elucidating the highly promising and appealing practical potential of finger pose interaction.

{
	\bibliographystyle{IEEEtran}
	\bibliography{egbib}{}
}

\begin{IEEEbiography}[{\includegraphics[width=1in,height=1.25in,clip,keepaspectratio]{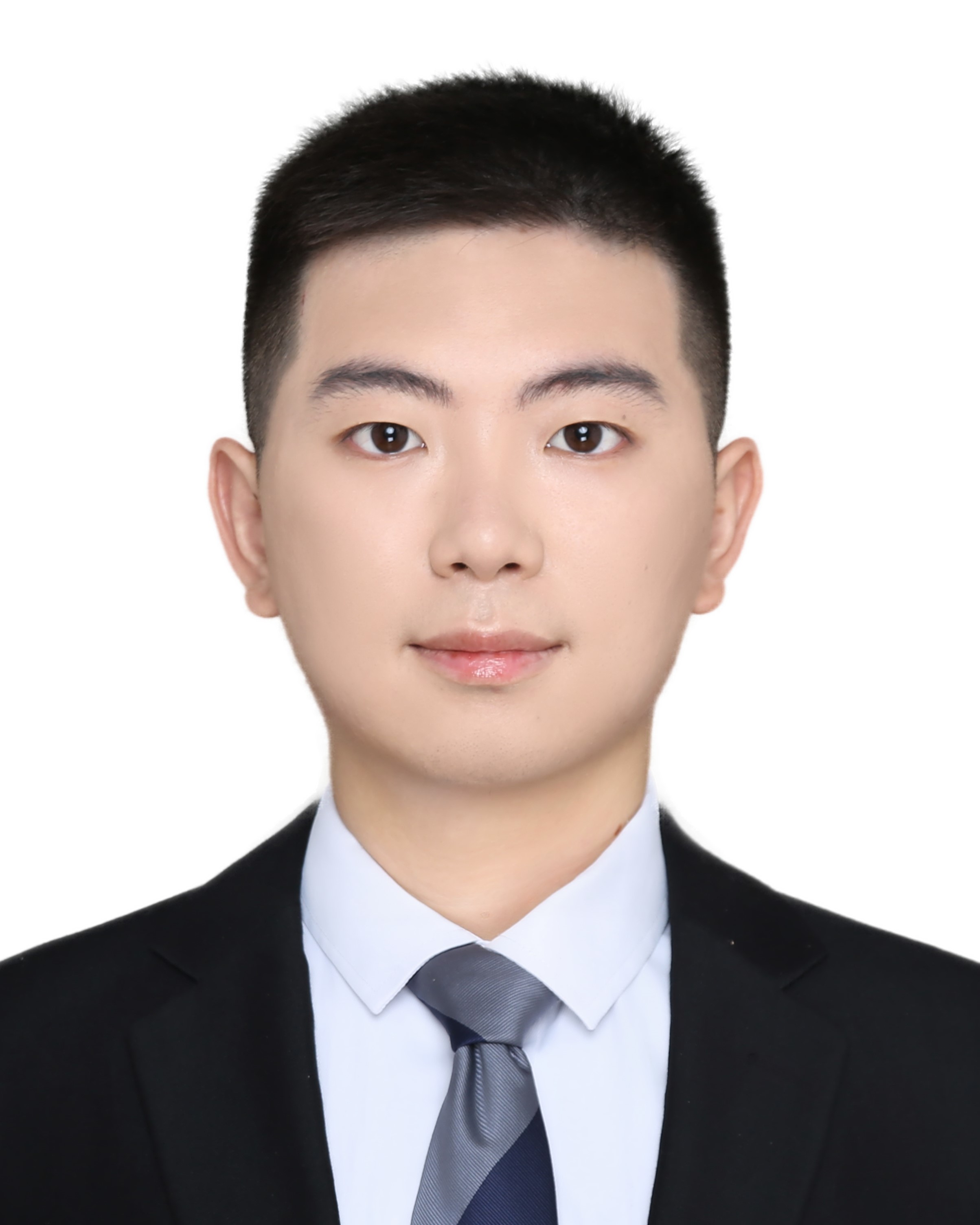}}]{Xiongjun Guan} received the B.Eng. degree in Department of Automation from Tsinghua University, Beijing, China, in 2021. He is currently pursuing the Ph.D. degree with the Department of Automation, Tsinghua University, under the supervision of Prof. Jianjiang Feng. His research interests include computer vision, pattern recognition, and human–computer interaction.
\end{IEEEbiography}

\begin{IEEEbiography}[{\includegraphics[width=1in,height=1.25in,clip,keepaspectratio]{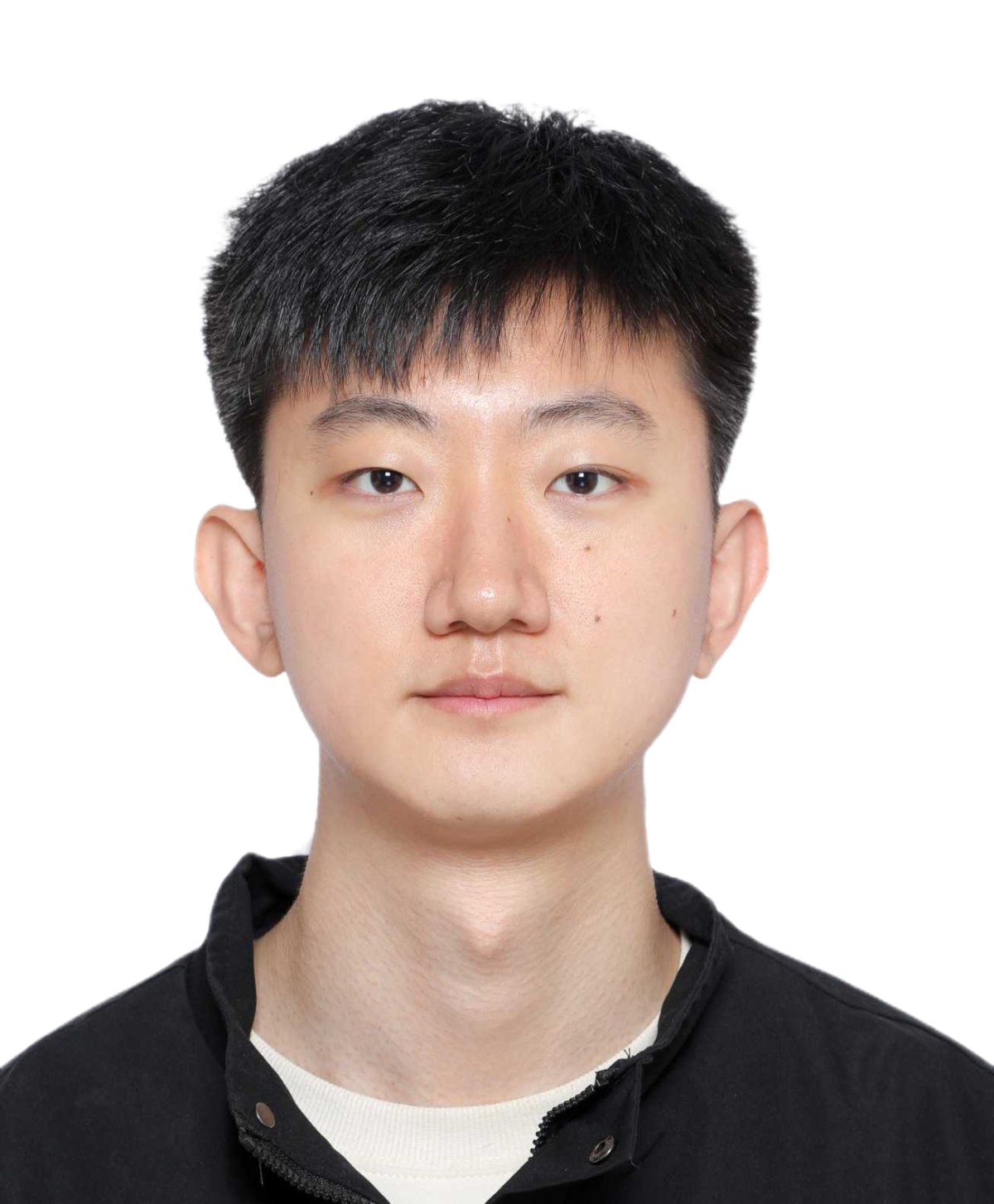}}]{Zhiyu Pan} received his Bachelor of Engineering (BEng) degree in Electronic Science and Technology from Beijing Institute of Technology, China, in 2020. He is currently pursuing a Ph.D. degree in the Department of Automation at Tsinghua University. His research interests include biometrics, human action analysis, and computer vision. Specifically, his current work focuses on fingerprint recognition, multi-modal learning, and related areas.
\end{IEEEbiography}

\begin{IEEEbiography}[{\includegraphics[width=1in,height=1.25in,clip,keepaspectratio]{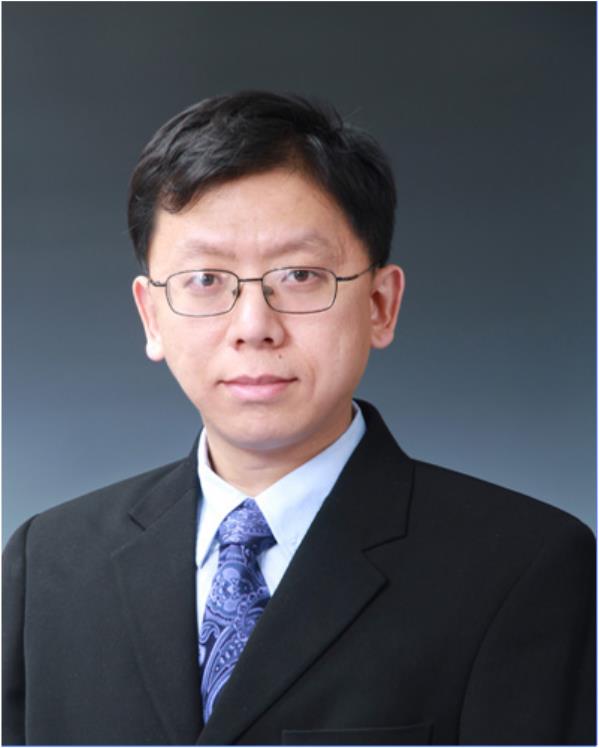}}]{Jianjiang Feng} (Member, IEEE) received the B.Eng. and Ph.D. degrees from the School of Telecommunication Engineering, Beijing University of Posts and Telecommunications, China, in 2000 and 2007, respectively. From 2008 to 2009, he was a Post-Doctoral Researcher with the PRIP Laboratory, Michigan State University. He is currently an Associate Professor with the Department of Automation, Tsinghua University, Beijing. His research interests include fingerprint recognition and computer vision.
\end{IEEEbiography}

\begin{IEEEbiography}[{\includegraphics[width=1in,height=1.25in,clip,keepaspectratio]{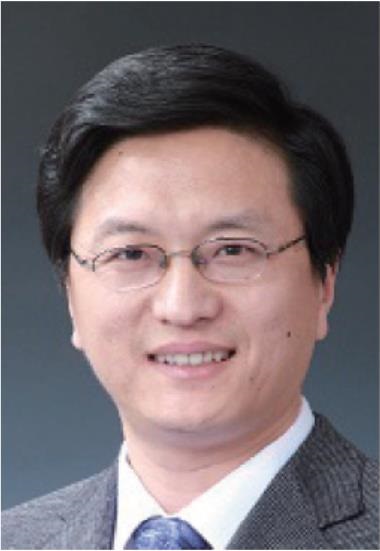}}]{Jie Zhou}  (Fellow, IEEE) received the BS and MS degrees both from the Department of Mathematics, Nankai University, Tianjin, China, in 1990 and 1992, respectively, and the PhD degree from the Institute of Pattern Recognition and Artificial Intelligence, Huazhong University of Science and Technology (HUST), Wuhan, China, in 1995. From then to 1997, he served as a postdoctoral fellow in the Department of Automation, Tsinghua University, Beijing, China. Since 2003, he has been a full professor in the Department of Automation, Tsinghua University. His research interests include computer vision, pattern recognition, and image processing. In recent years, he has authored more than 300 papers in peerreviewed journals and conferences. Among them, more than 100 papers have been published in top journals and conferences such as IEEE Transactions on Pattern Analysis and Machine Intelligence, IEEE Transactions on Image Processing, and CVPR. He is an associate editor for IEEE Transactions on Pattern Analysis and Machine Intelligence and two other journals. He received the National Outstanding Youth Foundation of China Award. He is an IAPR Fellow.
\end{IEEEbiography}

\vfill

\end{document}